\documentclass{article}
\usepackage[final]{neurips_2022}

\newcommand{\datamux}{DataMUX} 
\newcommand{\tmux}{T-MUX} 

\newcommand{\R}{\mathbb{R}}
\newcommand{\vm}[1]{{#1}}

\newcommand{\vmn}[1]{{#1}}

\renewcommand{\paragraph}[1] {\vspace{0.01in} \noindent {\bf #1 }}

\usepackage[utf8]{inputenc} % allow utf-8 input
\usepackage[T1]{fontenc}    % use 8-bit T1 fonts
\usepackage{hyperref}       % hyperlinks
\usepackage{url}            % simple URL typesetting
\usepackage{booktabs}       % professional-quality tables
\usepackage{amsfonts}       % blackboard math symbols
\usepackage{nicefrac}       % compact symbols for 1/2, etc.
\usepackage{microtype}      % microtypography
\usepackage{xcolor}         % colors
\usepackage{graphicx}
\usepackage{amsmath}
\usepackage{amssymb, bm}
\usepackage[labelfont=bf,font=small]{caption}
\usepackage{soul}
\usepackage{subcaption}
\title{DataMUX: Data Multiplexing for Neural Networks}

\author{%
  Vishvak Murahari\ \\
  Department of Computer Science\\
  Princeton University\\
  \texttt{murahari@princeton.edu} \\
   \And
  Carlos E. Jimenez\ \\
  Department of Computer Science\\
  Princeton University\\
  \texttt{carlosej@princeton.edu} \\
   \And
  Runzhe Yang\ \\
  Department of Computer Science\\
  Princeton University\\
  \texttt{runzhey@princeton.edu} \\
   \And
  Karthik Narasimhan\ \\
  Department of Computer Science\\
  Princeton University\\
  \texttt{karthikn@princeton.edu} \\
}

\begin{document}
\maketitle
% \begin{abstract}
% We introduce \emph{data multiplexing} (\datamux{}) -- a novel method for enabling neural networks to simultaneously process multiple inputs in a single inference pass. In contrast to techniques like batching, \datamux{} combines inputs into a single instance and requires only a small amount of additional memory. Our approach uses two key components -- a multiplexing layer that maps multiple input instances into different spaces before combining them to create a single `mixed' representation which is processed by the network, and a demultiplexing layer that converts the network's output back into independent representations which are then used to produce predictions for each input. We demonstrate the viability of \datamux{} on multiple architectures (Transformers, MLPs and CNNs), and for seven different tasks spanning sentence classification, named entity recognition and image classification. Our results show that \datamux{} is effective on multiplexing up to 40 input instances for a Transformer-based architecture, while sustaining a minimal drop in task performance (\todo{XYZ\%}) and increasing model inference speed by over 1,800\%. We also provide a theoretical construction for multiplexing in self-attention networks and perform ablations to study the effect of various design choices in \datamux{}.\footnote{Code and data are provided in supplementary material.}
% \end{abstract}

\begin{abstract}
In this paper, we introduce \emph{data multiplexing} (\datamux{}), a technique that enables deep neural networks to process multiple inputs simultaneously using a single compact representation. \datamux{} demonstrates that neural networks are  capable of generating accurate predictions over \emph{mixtures} of inputs, resulting in increased inference throughput with minimal extra memory requirements. Our approach uses two key components -- 1) a \textit{multiplexing} layer that performs a fixed linear transformation to each input before combining them to create a `mixed' representation of the same size as a single input, which is then processed by the base network, and 2) a \textit{demultiplexing} layer that converts the base network's output back into independent representations before producing predictions for each input. We show the viability of \datamux{} for different architectures (Transformers, and to a \vmn{much} lesser extent MLPs and CNNs) across six different tasks spanning sentence classification, named entity recognition and image classification. For instance, \datamux{} for Transformers can multiplex up to 20x/40x inputs, achieving \vmn{up to} 11x/18x increase in inference throughput with absolute performance drops of $<2\%$ and $<4\%$ respectively \vmn{compared to a vanilla Transformer} on MNLI, a natural language inference task. We also provide a theoretical construction for multiplexing in self-attention networks and analyze the effect of various design elements in \datamux{}.
\footnote{Code is available at \href{https://github.com/princeton-nlp/DataMUX}{https://github.com/princeton-nlp/DataMUX}}
\end{abstract}

\section{Introduction}
\label{sec:intro}

Deep neural networks are very effective at modeling complex functions from real-valued vector inputs to vector outputs.
However, recent studies have hinted that modern networks are vastly overparameterized~\cite{kaplan2020scaling,allen2019learning} and only require a fraction of their parameters for capturing many functions~\cite{frankle2018lottery,frankle2020linear}. This raises a natural question: \textit{if networks contain greater processing capacity than necessary, could it be possible for them to model a function over multiple inputs simultaneously, similar to how radio channels share bandwidth to carry multiple information streams at once?} Such a capability could enable significant improvements in efficiency and \vm{inference} throughput without requiring substantial increases in model size or computation.

\begin{figure}
    \centering
    \includegraphics[width=0.96\linewidth]{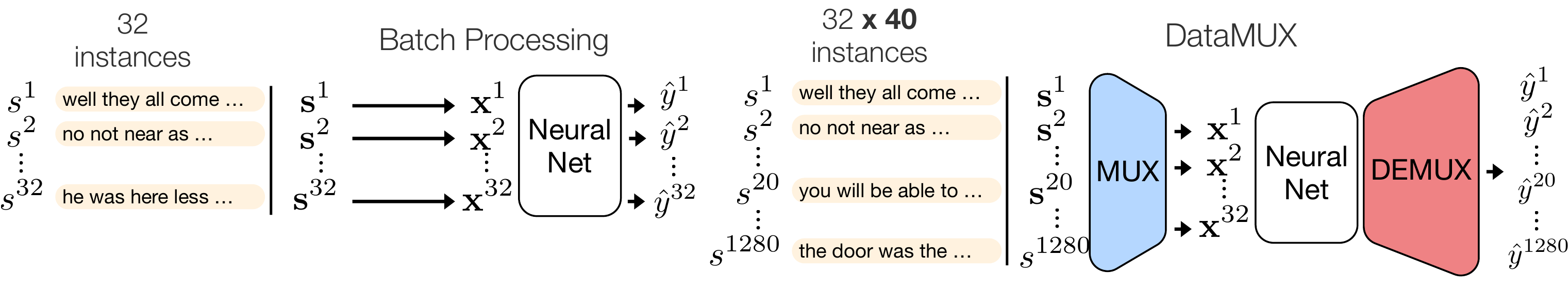}
    \caption{Schematic illustrating our proposed \datamux{} technique in comparison to traditional minibatch processing in neural networks. Here, \datamux{} uses a multiplexing layer (MUX), to multiplex $40$ instances into a single representation, passing only $32$ inputs to the neural network. The demultiplexing layer (DEMUX) uses the output to generate predictions for all $1280$ instances.}
    \label{fig:teaser}
\end{figure}

% In this paper, we demonstrate that neural networks can in fact be trained to produce multiple simultaneous predictions for a \textit{mixture} of input instances.
In this paper, we propose Data Multiplexing (\datamux{}), a technique that enables neural networks to process multiple inputs simultaneously as a single ``mixed'' representation of the same size as a single input, during both training and inference (Figure~\ref{fig:teaser}). This is complementary to and different from minibatch processing that adds an extra batch dimension to the input, which in effect increases computational and memory requirements. Due to the use of a single combined representation, \datamux{} can increase network inference throughput with minimal time and memory overhead.

%%%%%%
The key challenge for \datamux{} lies in effectively compressing a mixture of signals so that it can be processed by a neural network without sacrificing task performance. To this end, our procedure involves the addition of a multiplexing layer and a demultiplexing layer to both ends of a vanilla neural network. The multiplexing layer applies a fixed linear transformation to each input before combining them to create a single ``mixed'' representation. This is then fed into the base network, which processes it to obtain a ``mixed'' vector representation. The demultiplexing layer then converts this vector back into a set of vector representations corresponding to each original input, which is then used to make final predictions for each instance. Our multiplexing and demultiplexing layers are end-to-end differentiable, which allows us to train the entire model jointly through standard gradient descent methods. \vm{We find that training time to convergence increases almost linearly with the number of multiplexed in
stances as training gets more challenging since the model has to make multiple simultaneous predictions.} In order to train multiplexed models effectively, we also introduce a special token retrieval task as a pre-training objective. 

We demonstrate the effectiveness of our \datamux{} approach using three different architectures (Transformers \cite{NIPS2017_3f5ee243}, MLPs, and CNNs) on six tasks spanning sentence classification, named entity recognition and image classification. Our results demonstrate that \datamux{} models can successfully process between 2-40 input instances simultaneously, with minimal losses in task performance. For instance, on the sentiment analysis dataset SST-2, our multiplexed Transformer (\tmux{}) can handle 40x inputs with only a $2\%$ drop in accuracy compared to a vanilla Transformer. \vm{To our best knowledge, this is the first demonstration of large-scale data multiplexing in Transformers. While our results on CNNs and MLPs for image classification are not as strong as those on Transformers, we present them to demonstrate the generality of our method and believe that better multiplexing methods may yet improve them in the future.} 
% We also observe similar trends for MLPs and CNNs, albeit with less multiplexing capabilities in terms of the number of inputs. 

We further perform theoretical and empirical analyses to better understand our results. First, we show that it is possible to theoretically construct self-attention networks that can simultaneously process inputs in $N$ independent subspaces. Next, we perform several analyses of various multiplexing strategies and the effect of attention heads and model size. Finally, we also measure throughput statistics to demonstrate the efficiency gains provided by \datamux{} (up to 18x inference speedup).  We believe multiplexing holds great promise in improving the efficiency and versatility of neural networks and hope that future research can develop improved strategies for \datamux{} and expand its applicability to more architectures and tasks.

\section{Related Work}
\label{sec:related_work}

\paragraph{Multiplexing in deep learning}
While multiplexing is a standard idea in signal processing ~\cite{rabiner1975theory}, its use in deep learning has been limited. 
% However, the general idea of re-using neural network parameters to perform more than a single task has been studied under different settings.
 \citet{Lu2020MUXConvIM} develop MUXConv, a layer that multiplexes both spatial and channel information for convolutional neural networks (CNNs). Instead of performing \textit{data multiplexing}, these models multiplex different features within the network and only process one input at a time.
%  This helps them develop more compact networks with better utilization of parameters, but does not provide any data parallel capabilities like our approach. 
\textit{mixup} \cite{zhang2018mixup} proposes a training scheme where networks are trained with a convex combination of instances to predict the instances' label distributions from this combined representation. They find that this scheme provides implicit regularization of models and improves generalization during inference. Unlike our work however, \textit{mixup} does not preserve the order of the combined instances and therefore mixed instances may not be used during inference. Another line of work uses multiple-input multiple-output training  to obtain multiple sub-networks within a neural network for a relatively small number of inputs \cite{Havasi21, Rame21, Soflaei20}. These approaches aim to improve single-instance performance through ensembling sub-networks' predictions. In contrast, we are motivated by achieving optimal performance for prediction on multiple inputs and improving model throughput during inference.

\paragraph{Multiplexing in the brain}
On the biological side, several studies hint at the capability of the brain to multiplex and transmit tremendous amounts of information with a single neural code and process them in parallel ~\cite{blumhagen2011neuronal,akam2014oscillatory,pirschel2016multiplexed,hong2016multiplexed,friedrich2004multiplexing}. Recently, \citet{lankarany2019differentially} confirmed that neurons in the somatosensory cortex multiplex different types of features (low-contrast, high-contrast) using their timing and rate of synchronous and asynchronous spiking. \citet{rezaei2021synchrony} further provide a computational model to explain multiplexing with an ensemble of homogeneous cortical neurons. These studies increasingly demonstrate that data multiplexing is a key feature of \emph{natural} neural networks.

\paragraph{Over-paramaterization in neural networks}
Our work also relates to the over-parameterization of deep neural networks that has been investigated in several pieces of prior work~\cite{allen2019convergence,allen2019learning,radhakrishnan2020overparameterized}. The lottery ticket hypothesis~\cite{frankle2018lottery,frankle2020linear,malach2020proving} demonstrated that only a fraction of neurons in a trained deep model are sufficient to capture a required function.  Neural architecture search~\cite{zoph2016neural,liu2018progressive,elsken2019neural} aims to balance performance with parameter efficiency, by discovering the most optimal architectures for each task. There have also been several efforts at leveraging this over-parameterization effect for multi-task learning~\cite{hu2021unit}. \citet{NEURIPS2019_4c7a167b} propose a method to superpose many models for different tasks into a single neural network, and show that it helps mitigate catastrophic forgetting. \citet{NEURIPS2020_ad1f8bb9} is another work along the same lines which uses a base network, along with specialized sub-networks for different tasks. 
Similar to these paradigms, our work possibly takes advantage of the over-parameterization existing in deep neural networks. However, to our knowledge, we are the first to utilize \emph{data} multiplexing to simultaneously process multiple inputs during both training and inference, and are able to show that a fixed set of parameters can process a mixture of inputs (even up to 40) with minimal loss in accuracy.

%TODO:
%Connections with:
%- Lottery ticket hypothesis
%- Small transformer models; studies on overparameterization?
%- Pathways (multi-task models)
%- architecture search
\section{Method}
\label{sec:method}

% \subsection{Overview}

% Should we introduce the task of data multiplexing using the Cartesian product notation? (X^n -> Y^n) Perhaps in an earlier section?

\begin{figure}[ht]
\centering
  \begin{minipage}[t]{0.49\textwidth}
    \includegraphics[width=0.97\textwidth]{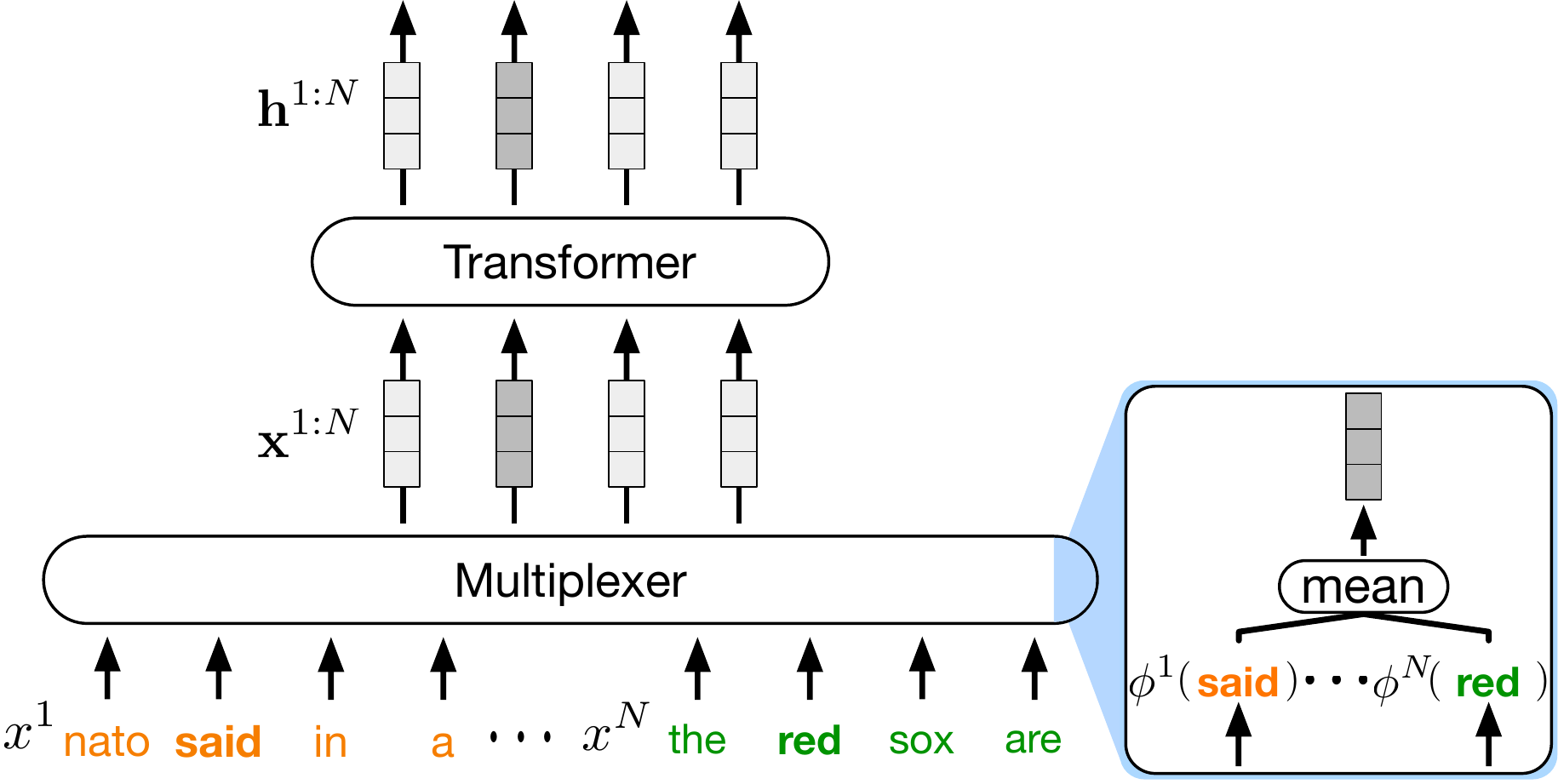}
    % \caption{\vm{\textbf{Multiplexing + Encoding: }}Given a tuple of $N$ sentences ($x^1, x^2, \dots, x^N$), each of length $L$, we first apply a multiplexing operation which performs a transformation $\phi^i$ on the embeddings of each sequence $x^i$, such that the same transformation is applied to every token in a sequence $x^i$. The multiplexing operation then aggregates the sequences by averaging over each position, generating a single combined sequence $\mathbf{x}^{1:N}\in\mathbb{R}^{L\times d}$, for embedding size $d$, which will be passed on to the central Transformer model.}
    % \label{fig:tmux_schematic_mux}
  \end{minipage}
  \hspace{5pt}
  \begin{minipage}[t]{0.48\textwidth}
    \includegraphics[width=0.97\textwidth]{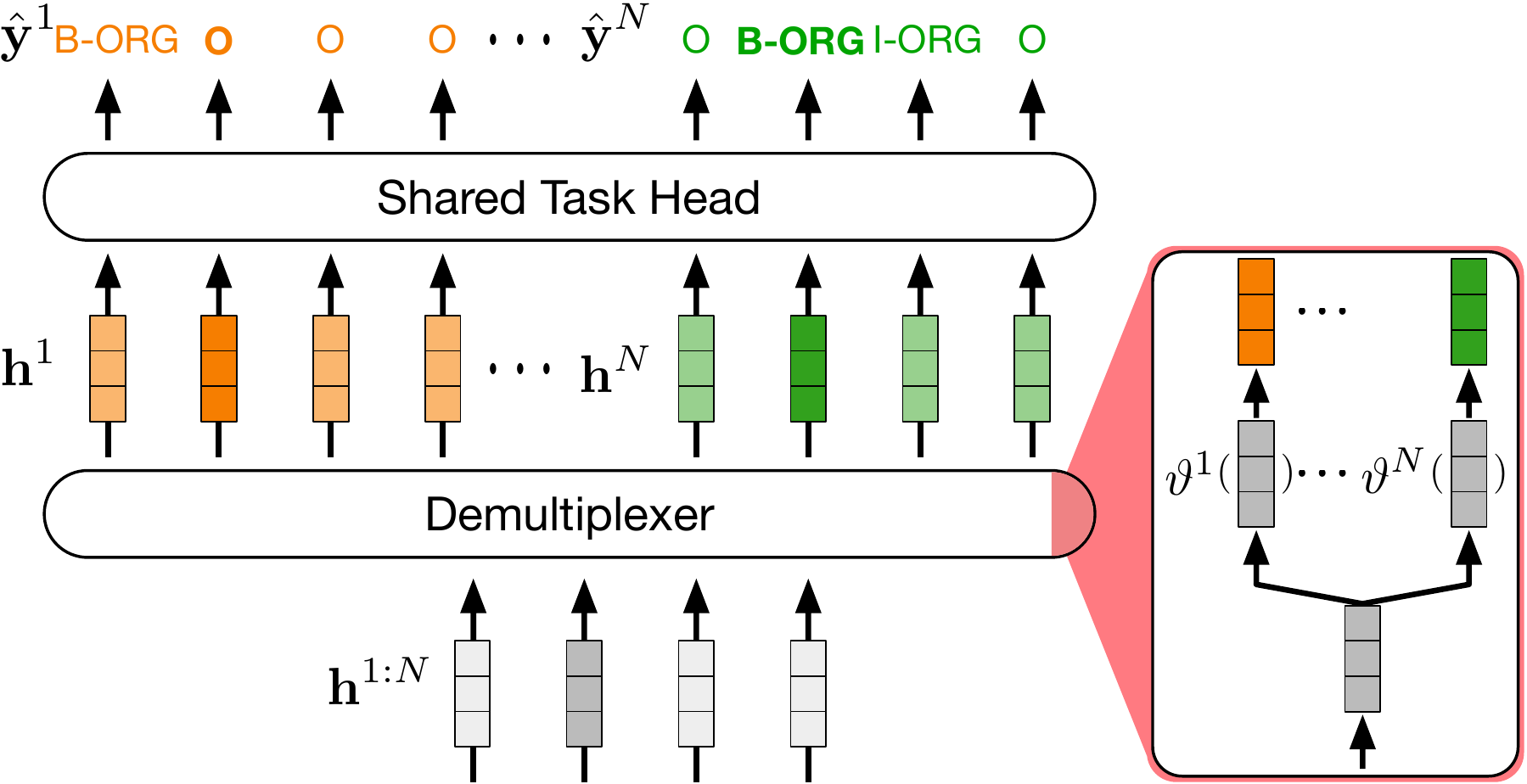}
    % \caption{\vm{\textbf{Demultiplexing: }}After processing, we perform a demultiplexing operation to the Transformer model's output $\mathbf{h}^{1:N}\in\mathbb{R}^{L\times d}$, to generate hidden representations $\mathbf{h}^1, \mathbf{h}^2, \dots, \mathbf{h}^N$, corresponding to inputs $x^1, x^2, x^N$ respectively. We finally use these hidden representations to generate predictions for a particular task (e.g. named entity recognition (NER)) using a shared task prediction head.}
    % \label{fig:tmux_schematic_demux}
  \end{minipage}
 \caption{\datamux{} for Transformers: \textbf{(Left)} Given a tuple of $N$ sentences ($x^1, x^2, \dots, x^N$), each of length $L$, we first apply a \textit{multiplexing} operation which performs a transformation $\phi^i$ on the embeddings of each sequence $x^i$, such that the same transformation is applied to every token in a sequence $x^i$. The multiplexing operation then aggregates the sequences by averaging over each position, generating a single combined sequence $\mathbf{x}^{1:N}\in\mathbb{R}^{L\times d}$, for embedding size $d$, which will be passed on to the central Transformer model. \textbf{(Right)} After processing, we perform a \textit{demultiplexing} operation to the Transformer model's output $\mathbf{h}^{1:N}\in\mathbb{R}^{L\times d}$, to generate hidden representations $\mathbf{h}^1, \mathbf{h}^2, \dots, \mathbf{h}^N$, corresponding to inputs $x^1, x^2, x^N$ respectively. We finally use these hidden representations to generate predictions for a particular task (e.g. named entity recognition (NER)) using a shared task prediction head.}
 \label{fig:tmux_schematic}
\end{figure}

Our primary goal is to process a set of input instances \emph{simultaneously} over a shared neural network (multiplexing) with minimal overhead during inference. To this end, we design \datamux{} to consist of three components: a multiplexer module to combine the multiple input instances into a superposed representation, a neural network backbone to process mixed representations, and a demultiplexing module to disentangle the processed representations for individual prediction. A schematic illustrating the flow of representations is shown in Figure~\ref{fig:tmux_schematic}. We detail the general requirements for each component below, as well as specific implementation details used in our experiments.

\subsection{Multiplexing}

The multiplexer module, denoted $\Phi$, combines a tuple of inputs, either images or sentences from a batch, $(\mathbf{x}^1,\dots,  \mathbf{x}^N)$, for $\mathbf{x}^i \in \R^d$, into a more compact representation $\mathbf{x}^{1:N} \in \R^d$ in an order-dependent way, which enables effective demultiplexing after processing, as well as distinguishing intra-sequence interactions in the case of sequenced inputs (e.g. token sequences). Towards this end, for each input $\mathbf{x}^i$ with index $i\in [1, N]$ of the input tuple, the multiplexer module performs a transformation $\phi^i$ ($\R^d$ $\mapsto$ $\R^d$), on the instance before finally averaging all inputs into a single multiplexed representation as:
% $$
%     \mathbf{x}^{1:N}  = \Phi(\mathbf{x}^1,\dots,  \mathbf{x}^N) = \frac{1}{N}\sum_{i=1}^{N} \phi^i(\mathbf{x}^i)
% $$
\setlength{\abovedisplayskip}{3pt}
\setlength{\belowdisplayskip}{3pt}
\begin{equation}
    \mathbf{x}^{1:N}  = \Phi(\mathbf{x}^1,\dots,  \mathbf{x}^N) = \frac{1}{N}\sum_{i=1}^{N} \phi^i(\mathbf{x}^i).
    \label{eq:combined-rep}
\end{equation}
For sequenced inputs (e.g. token sequences), we combine $N$ sequences by multiplexing token-wise. That is for inputs of the form $\mathbf{x}^i = \{\mathbf{w}^i_j\}_{j\in[1, L]}$, where $\mathbf{w}^i_j\in\R^d$ is a token's input vector representation, this operation uses the same transformation $\phi^i$ for each token in the sequence before averaging over each position across indices, such that $\mathbf{x}^{1:N} = \{\mathbf{w}^{1:N}_{j}\}_{j\in[1, L]}$ where $\mathbf{w}_j^{1:N}\in\R^d$ is a multiplexed representation of $N$ tokens at position $j$, i.e., $\mathbf{x}^{1:N} = \{\Phi(\mathbf{w}^{1}_j, \dots, \mathbf{w}^{N}_j)\}_{j\in [1,L]}$.

% This is equivalently denoted:

% $$
%     \mathbf{x}^{1:N}  = (\Phi(\mathbf{w}^1_1, \dots, \mathbf{w}^N_1),\dots, \Phi(\mathbf{w}^1_L, \dots, \mathbf{w}^N_L))
% $$

For $\phi^i$, we experiment with using either (1) a linear projection with a random fixed orthogonal matrix (denoted 
``Ortho'') or (2) the Hadamard product with a fixed Gaussian random vector (denoted ``Hadmard'', equivalent to a linear map using a diagonal matrix in our case). We hope these transformations map instances at different indices into distinguishable regions and consequently reduce interference between their representations. Finally, this multiplexed representation, $\mathbf{x}^{1:N}$, is used as input to the neural network backbone, which is architecturally unchanged.

\subsection{Demultiplexing}
\label{sec:demuxing}

\setlength{\abovedisplayskip}{3pt}
\setlength{\belowdisplayskip}{3pt}
The output of the neural network backbone will be a multiplexed hidden representation $\mathbf{h}^{1:N}$ of the input $\mathbf{x}^{1:N}$. To make a prediction for each input, one can explicitly disentangle $\mathbf{h}^{1:N}$ into $N$ individual hidden representations, $\mathbf{h}^{1},\dots, \mathbf{h}^{N}$, respectively. We first obtain each $\mathbf{h}^i$ with a demultiplexing function $\vartheta^i$, i.e.,
\begin{equation}
    \mathbf{h}^i = \vartheta^{i}(\mathbf{h}^{1:N}),~ \forall i\in[1,\dots, N].
    \label{eq:demux-rep}
\end{equation}

% $$
% (\mathbf{h}^1,\dots,  \mathbf{h}^N) = \vartheta(\mathbf{h}^{1:N}) = (\vartheta^1(\mathbf{h}^{1:N}),\dots, \vartheta^{N}(\mathbf{h}^{1:N}))
% $$

For sequenced input, demultiplexing is done position-wise, i.e., $\mathbf{h}^i_j = \vartheta^i(\mathbf{h}^{1:N}_j)$ for each position $j$.

% $$
%     \vartheta^{\text{seq}}(\mathbf{h}^{1:N})  = (\vartheta(\mathbf{h}^{1:N}_1),\dots, \vartheta(\mathbf{h}^{1:n}_L))
% $$

Finally, predictions are made using a shared task head on each inputs' respective individual hidden representation to prevent a  substantial increase in the number of parameters and improve training efficiency. We use two alternatives for the demultiplexing function $\vartheta^i$:

% \subsubsection{Generating individual hidden representations} 

% \textbf{MLP Demuxing} and \textbf{Index Embeddings}. 
\textbf{1. MLP Demuxing} This strategy employs $N$ MLPs to generate each indices' hidden representation as $\mathbf{h}^i = \textsc{mlp}^i(\mathbf{h}^{1:N})$. We use this method for both NLP and vision tasks. Although this method is conceptually simple, it adds learnable parameters proportional to $N$.\\
% which could potentially lead to optimization issues for large $N$.\\

% \phantomsubsection
\label{sec:index_embedding}
\textbf{2. Index Embeddings} We generate index embeddings $\mathbf{p}^i$, which are then concatenated to $\mathbf{h}^{1:N}_j$, and transformed by a shared multi-layer network to generate each individual hidden representation, i.e.,  $\mathbf{h}^i_j = \textsc{mlp}^{\text{shared}}( \mathbf{h}_j^{1:N}, \mathbf{p}^i)$. To generate the index embeddings $\mathbf{p}^i$, we add a sequence of $N$ special tokens, called the prefix, to the beginning of each sequence of the input tuple. For multiplexing with $N$ sequences, we add $N$ corresponding prefixes, denoted $\textit{prefix}^i$ for $i\in[1, N]$. Each $\textit{prefix}^i$ consists of a index token $\epsilon^{i}$ in it's $i$'th position while the remaining tokens are a special pad token $\epsilon^{\text{pad}}$. The prefix sequences then take on the following pattern:
    \begin{flalign*}
    &\textit{prefix}^1 \hspace{0.3em}= [\epsilon^1, \epsilon^{\text{pad}}, \ldots, \epsilon^{\text{pad}} ]
    &\textit{prefix}^2 \hspace{0.3em}= [\epsilon^{\text{pad}}, \epsilon^2, \epsilon^{\text{pad}}, \ldots, \epsilon^{\text{pad}} ]
    &\cdots
    &\textit{prefix}^N = [\epsilon^{\text{pad}}, \ldots, \epsilon^{\text{pad}}, \epsilon^N ]
    \end{flalign*}

We then prepend each sequence $x^i$ of the input sequence with the corresponding $\textit{prefix}^i$. The tuple of prepended sequences is then passed to the multiplexing module. When finally generating individual hidden representations, we use the corresponding hidden representation of each index token $\epsilon^i$ as the index embedding $\mathbf{p}^i$.

We use the Index Embeddings (~\ref{sec:index_embedding}) demultiplexing strategy  on language tasks for the Transformer architecture. The prefix tokens may implicitly enable the Transformer to do instance-specific computations when processing the multiplexed representation and further enable demultiplexing for large $N$.

\subsection{Retrieval warm-up for multiplexing}
\label{sec:retrieval_methods}
For Transformer models, we find that naively adding the multiplexing and demultiplexing layers to the model fail to converge. This is likely because the gradients for individual instances from the task loss get mixed up in the backward pass.
To overcome this, we propose \textit{Retrieval} warm-up -- a self-supervised pre-training task to promote the ability of \datamux{} models at distinguishing the order and the content of individual sequences in a multiplexed representation. This task consists of retrieving the correct tokens and order for each position and sequence of the input tuple (Figure~\ref{fig:retrieval_schematic}). Although we could add this loss for every sentence in each position, memory constraints force us to retrieve a token from one random sentence for each token position, yielding the following objective:
% $$
% \mathcal{L}_{\text{Retrieval}}(x^{1:N}) = \sum_{j} - \log P(w^{\text{rand}([1, N])}_j|\mathbf{h}^{\text{rand}([1, N]}_j)
% $$
\begin{equation}
    \mathcal{L}_{\text{Retrieval}}(x^{1:N}) = \sum_{j} - \log P(w^{I}_j|\mathbf{h}^{I}_j),
    \label{eq:retrieval-loss}
\end{equation}
where $\mathbf{h}^I_j$ is a demultiplexed hidden representation of the $j$-th token in a randomly selected sentence with the index $I\sim \mathcal U[1,N]$, generated using the methods described in Section~\ref{sec:demuxing}. \vm{We use this objective to optimize all the parameters of the end-to-end multiplexed model.}
For sequenced inputs, we find the retrieval auxiliary objective crucial to learning and report performance on the retrieval task in Section~\ref{sec:retrieval_results}.

% \paragraph{Using input prefix for conditional vectors}

% \subsubsection{MLP-based demultiplexing}

% Why retrieval of some variety is necessary.
% Should we avoid analogies to MLM? (80/10/10)

\section{Multiplexing for Transformers (\tmux{})}
\label{sec:transformers_mux_expts}

\subsection{Experimental setup}

\paragraph{Models}
We first evaluate the capabilities and limits of data multiplexing specifically for the Transformer architecture on a range of text classification tasks. We apply \datamux{} on a 12-layer Transformer based model with a hidden dimension size of 768 and 12 self-attention heads built on the Huggingface~\cite{Wolf2019HuggingFacesTS} framework, and refer to the resulting model as \tmux{}. We compare our \tmux{} models to 2 baselines: {\bf(B1)} A 12-layer 768 hidden dimension vanilla Transformer. {\bf(B2)} A 12-layer 768 hidden dimension Transformer pretrained using the retrieval task described in Section \ref{sec:retrieval_methods}. Though there is no multiplexing done for {\bf B2} (meaning this operation could be solved by simply copying input tokens to the output layer) we find that the retrieval pre-training produces differences in performance and we show this baseline for completeness.

We also apply \datamux{} to smaller models (see \hyperref[sec:small-model-datamux]{\textbf{A2}}) and compare with similar baselines.

\paragraph{Tasks}
We evaluate our models and the baselines on two types of text classification tasks: 

\textbf{1. Token-level classification:} This evaluates models' ability to perform token-level tasks on multiplexed inputs. This poses a particular challenge for data-multiplexing models since they must maintain a high level of individual token disentanglement while also producing representations capable of solving the task. We evaluate token-level classification on the CoNLL-2003 Named Entity Recognition (NER) task~\cite{Sang2003IntroductionTT}.

\textbf{2. Sentence-level classification:} We evaluate models on a subset of the sentence-level classification tasks found in the General Language Understanding Evaluation (GLUE) benchmark~\cite{wang2019glue}: the sentiment classification task SST-2~\cite{socher2013recursive}, the sentence similarity task QQP~\footnote{\href{https://data.quora.com/First-Quora-Dataset-Release-Question-Pairs}{https://data.quora.com/First-Quora-Dataset-Release-Question-Pairs}}, and the natural language inference tasks MNLI \cite{williams2018broad} and QNLI \cite{wang2019glue, rajpurkar2016squad}. By evaluating on a variety of sentence-level tasks, we can gain a better sense of the capabilities of data multiplexing neural networks on tasks that require aggregating contextual information. Similar to previous works, we prepend a [CLS] token to all sequences and learn a task head on top of the demultiplexed [CLS] token representation.

\paragraph{Auxiliary retrieval objective}
The \tmux{} models are all pre-trained using the retrieval warm-up on the Wikitext-103 dataset~\cite{Merity2017PointerSM} as described in Section~\ref{sec:retrieval_methods}. In addition, we also continue to use the retrieval task as an auxiliary objective during task training. The total loss is a combination of the task loss and retrieval loss (we use $\alpha = 0.1$ in our experiments):
% $$
% \mathcal{L} = (1-\alpha)\mathcal{L}_{\text{task}} + \alpha\mathcal{L}_{\text{Retrieval}}
% $$
\begin{equation}
    \mathcal{L} = (1-\alpha)\mathcal{L}_{\text{Task}} + \alpha\mathcal{L}_{\text{Retrieval}},
    \label{eq:total-loss}
\end{equation}

\subsection{Main results}

\begin{figure*}[t]
    \centering
\includegraphics[width=0.95\textwidth]{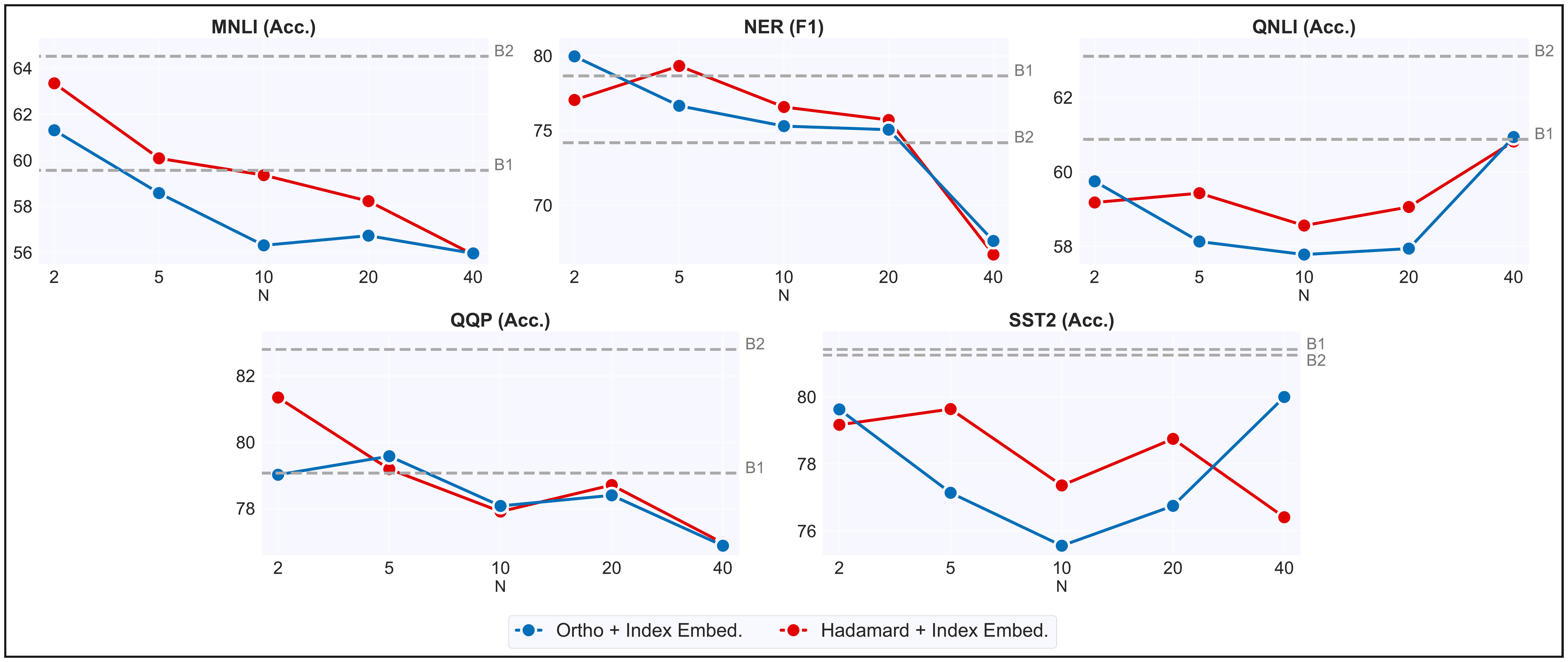}
    \caption{Multiplexing task evaluation for \tmux{}. Across tasks, we demonstrate multiplexing up to $40$ instances without significant drop in performance. Further, we show results here for Index Embedding demultiplexing as the MLP Demuxing method leads to optimization instability for the Transformer architecture. We provide results for the MLP Demuxing method in Appendix \ref{appendix:mlp_demux}.}
    \label{fig:large_results_chart}
\end{figure*}

\paragraph{(R1) Multiplexing leads to minimal drop in performance even for large $N$}
Figure~\ref{fig:large_results_chart} shows performance across four sentence classification tasks (MNLI, QNLI, QQP, SST2) and a token-level classification task (NER). We observe that it is possible to multiplex up to $40$ instances and maintain reasonable performance. For easier tasks like QQP, SST2 and QNLI, we observe that the drop in performance with increasing $N$ is insignificant while for more difficult tasks like MNLI and NER, there is a trade-off between performance and $N$, with performance dropping 10\%-15\% for $40$ instances. Since we've encountered unstable optimization when using MLP Demuxing, we only provide results using Index Embeddings demultiplexing. We find that the multiplexing strategy does not impact performance across different tasks, even slightly increasing for small values of $N$ (e.g. 2, 5). This increase may be attributable to implicit regularization à la \textit{mixup}. We also did not notice much variance between fine-tuning runs and plot variance for the "Hadamard + Index Embedding" setting in Section~\ref{appendix:variance}.

\begin{figure}
    \begin{subfigure}[ht]{0.34\linewidth}
        \centering
        \includegraphics[width=\columnwidth]{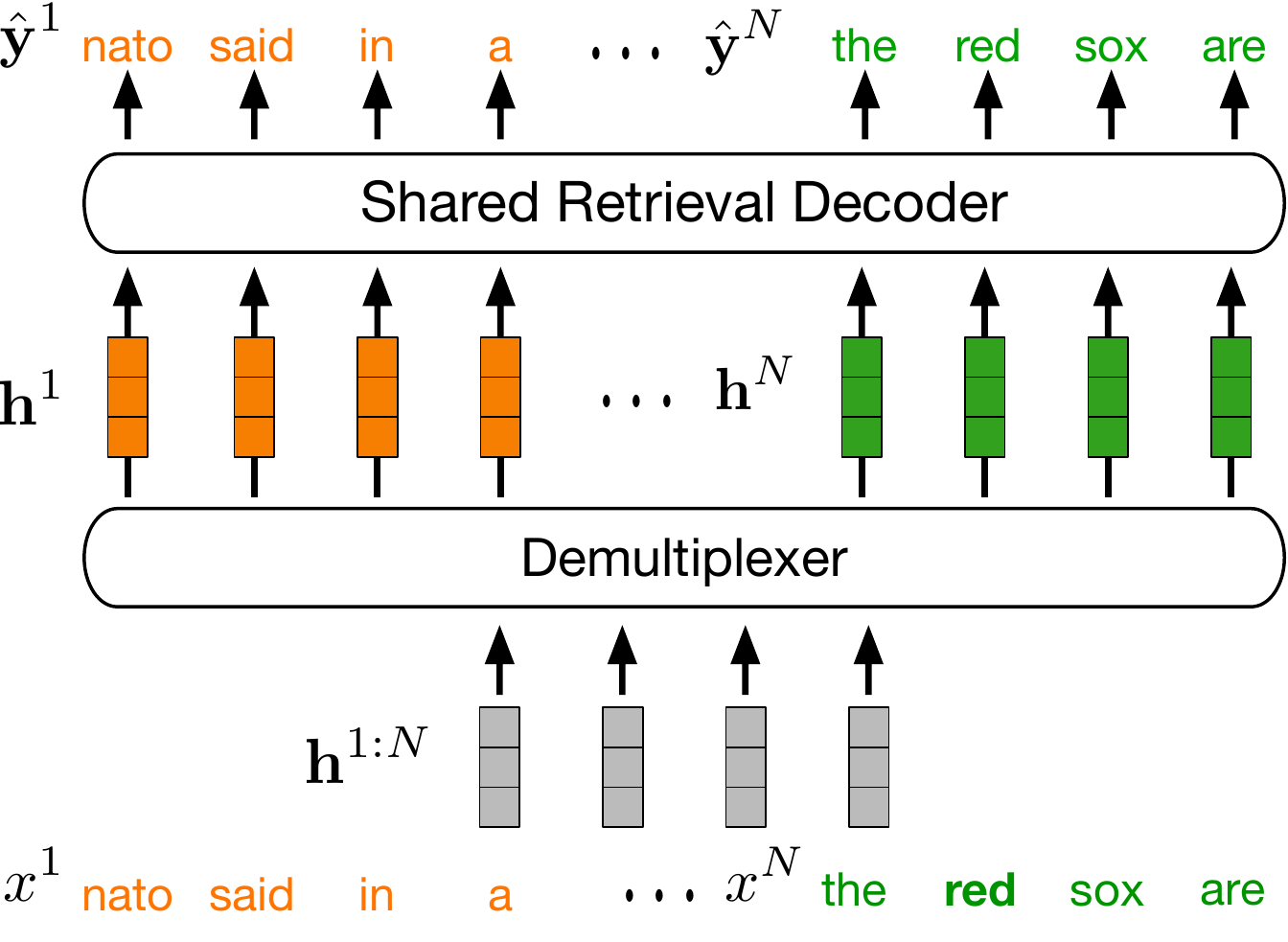}
     \phantomsubcaption
    \label{fig:retrieval_schematic}
    \end{subfigure}
    % \begin{subfigure}[ht]{0.48\linewidth}
    % \phantomsubcaption
    % \label{fig:retrieval_chart}
    % \includegraphics[width=0.8\linewidth]{charts/retrieval_accuracy.pdf}
    % % \caption{Accuracy for the retrieval warm-up task. Surprisingly, models are able to retrieve words with $100\%$ accuracy up to $20$ instances across most multiplexing and demultiplexing strategies.}
    % \end{subfigure}
    % \hspace{10pt}
  \begin{subfigure}[ht]{0.23\linewidth}
    \includegraphics[width=\linewidth]{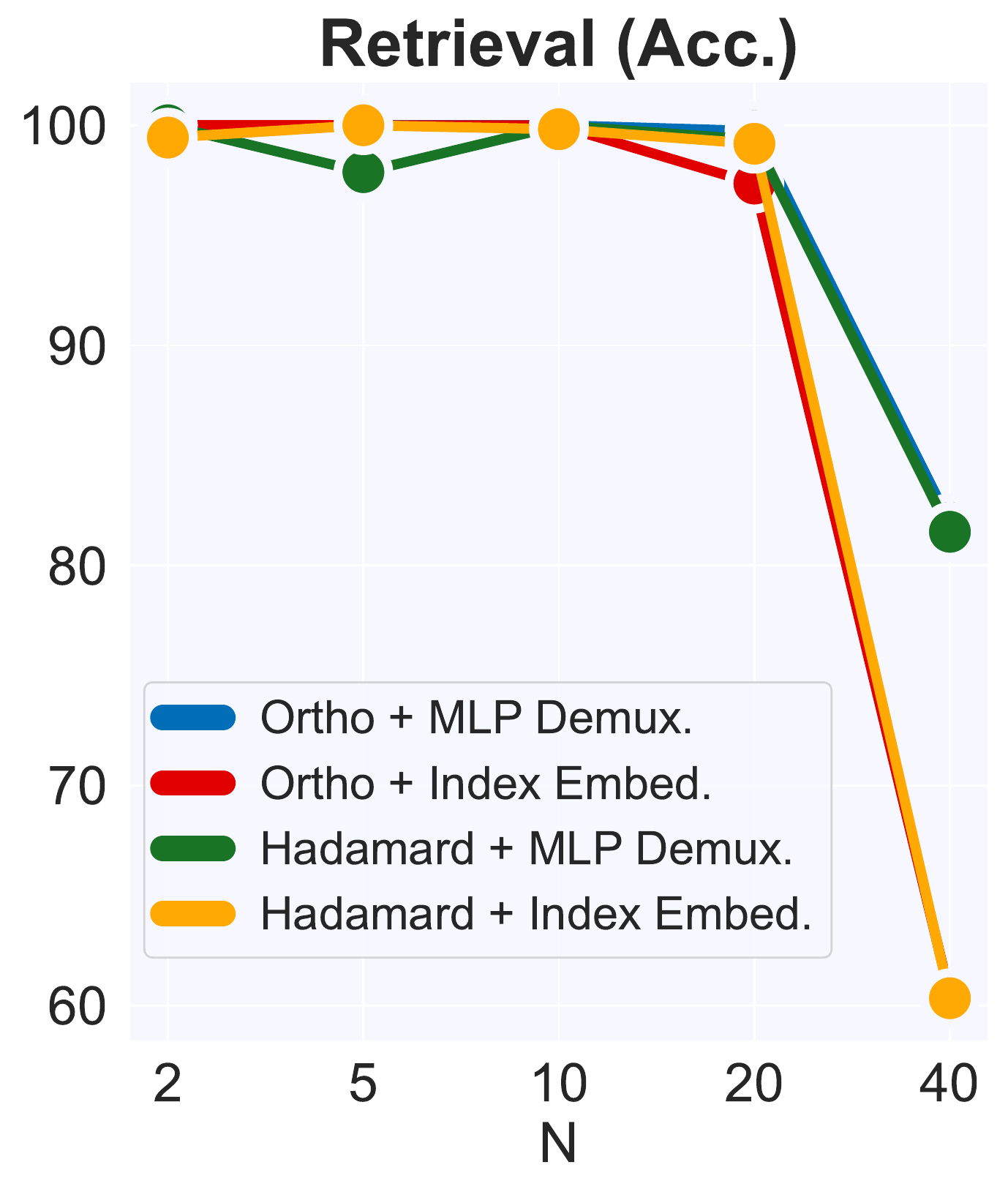}
    \centering
    \phantomsubcaption
    % \caption{Average test accuracy for multiplexing MLP and CNN for $N=\{1, 2, 4, 8, 16\}$ on the MNIST classification task for different multiplexing strategies. MLPs can be multiplexed up to $N=8$ without significant drop in performance while CNNs can only be multiplexed up to $4$ instances. \kn{maybe say (baseline) for ID models in legend}}
    \label{fig:retrieval_chart}
    \end{subfigure}
    \begin{subfigure}[ht]{0.43\linewidth}
        \centering
        \includegraphics[width=\columnwidth]{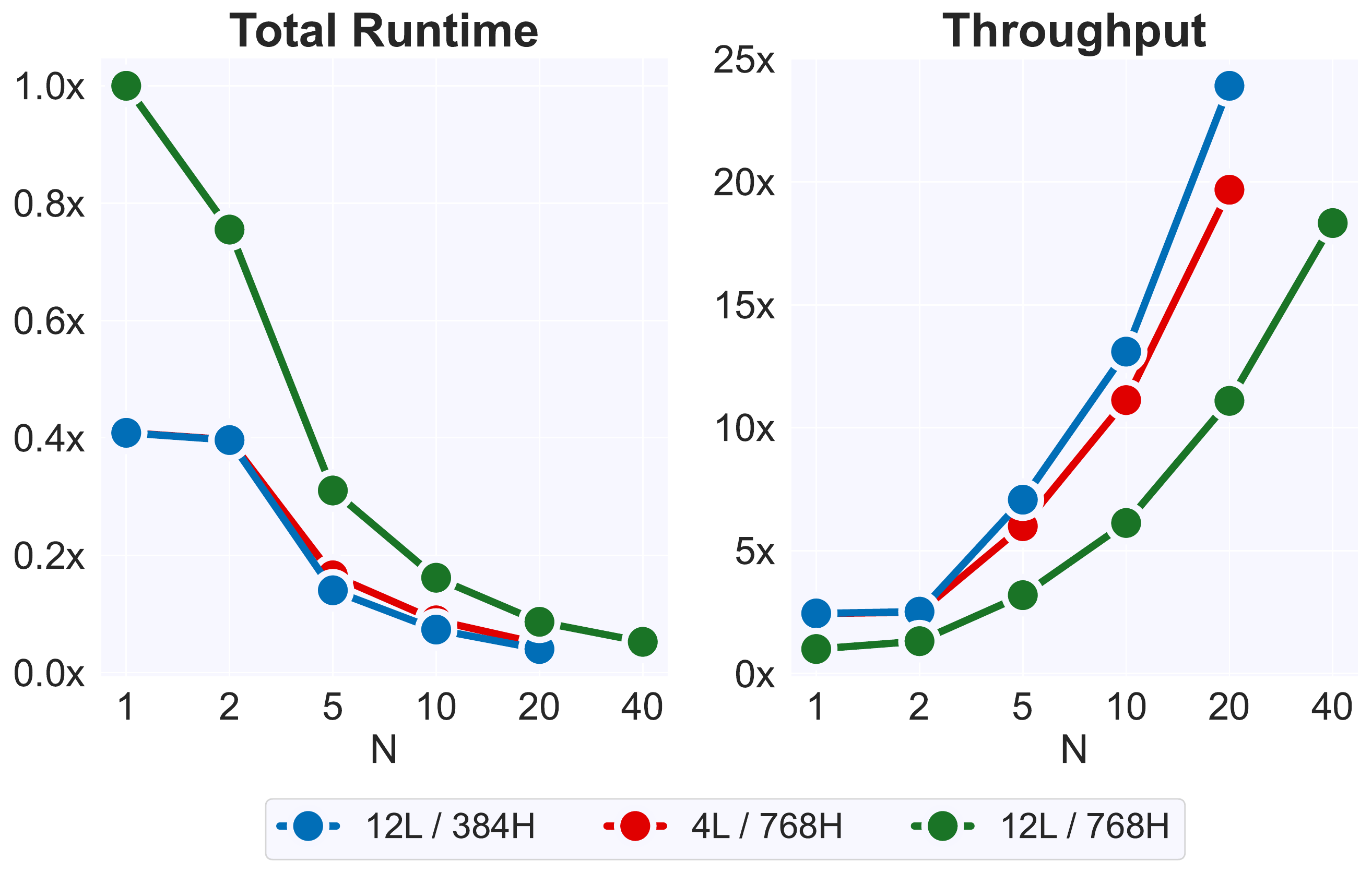}
        % \caption{Runtime and throughput numbers of \tmux{} models on 20k MNLI instances, normalized by a base model's performance without multiplexing. \tmux{} (12L/768H) can multiplex up to $40$ instances leading to $\sim$ $18$x speedup.}
     \phantomsubcaption
    \label{fig:efficiency_chart}
    \end{subfigure}
    \caption{\textbf{(Left a)} Given the Transformer's combined output representation $\mathbf{h}^{1:N}$ generated from the input $(x^1, x^2, \dots, x^N)$, the goal of the retrieval task is to retrieve the original tokens from $(x^1, x^2, \dots, x^N)$. Since $\mathbf{h}^{1:N}$ is order-dependent, retrieval requires distinguishing not just which tokens were originally input to a position $j\in[1, L]$, but also which sentence $i\in[1, N]$ each belonged to. \textbf{(Center b)} Accuracy for the retrieval warm-up task. Surprisingly, models are able to retrieve words with $100\%$ accuracy up to $20$ instances across most multiplexing and demultiplexing strategies. \textbf{(Right c)} Runtime and throughput numbers of \tmux{} models on 20K MNLI instances, normalized by a base model's performance without multiplexing. \tmux{} (12L/768H) can multiplex up to $40$ instances leading to $\sim$ $18$x speedup.
    \vspace{6pt}}
\end{figure}

\paragraph{(R2) Perfect multiplexing on the retrieval warm-up task for large $N$}
\label{sec:retrieval_results}
Figure ~\ref{fig:retrieval_chart} shows the test accuracy on the retrieval warm-up task described in Section~\ref{sec:retrieval_methods}. We first note that across different multiplexing and demultiplexing strategies, models have a retrieval accuracy of nearly $100\%$ for up to $20$ instances, demonstrating the surprising ability of \tmux{} to multiplex perfectly for large $N$. Note that this task does not require any aggregation of context across the sequence and thereby is much easier than sentence classification or token-level classification tasks. Therefore, performance on this warm-up task indicates a soft upper bound on the number of instances we can multiplex for sentence and token-level classification tasks given a particular multiplexing and demultiplexing method.
% \todo{Figure 4 comes before figure 3}
% \vm{Would it make sense to swap R1 and R2}

\paragraph{(R3) Throughput can be increased multi-fold}
\label{sec:throughput}
We measure throughput of our multiplexed model (Hadamard + Index Embed) across different number of instances by calculating inference speed for processing $\sim$20,000 samples on the MNLI dataset. We use four different batch sizes for all the configurations and take the max throughput (details in \ref{appendix:throughput_expt_details}).
 Figure~\ref{fig:efficiency_chart} shows that multiplexing increases throughput many folds ($18$x for $40$ instances, $11$x for $20$ instances). \tmux{} enables superior throughput as batch size can be effectively increased by a factor of $N$. We would expect the speedup to scale linearly with $N$, however a large $N$ corresponds to more prefix tokens which increases the sequence length. Therefore, having $40$ instances leads to almost a $20$x speedup as opposed to the expected $40$x. Future work can potentially improve this speedup by designing better multiplexing and demultiplexing strategies.

\subsection{Analysis}

\begin{figure}
  \centering
  \begin{subfigure}[ht]{0.48\linewidth}
    \includegraphics[width=0.95\linewidth]{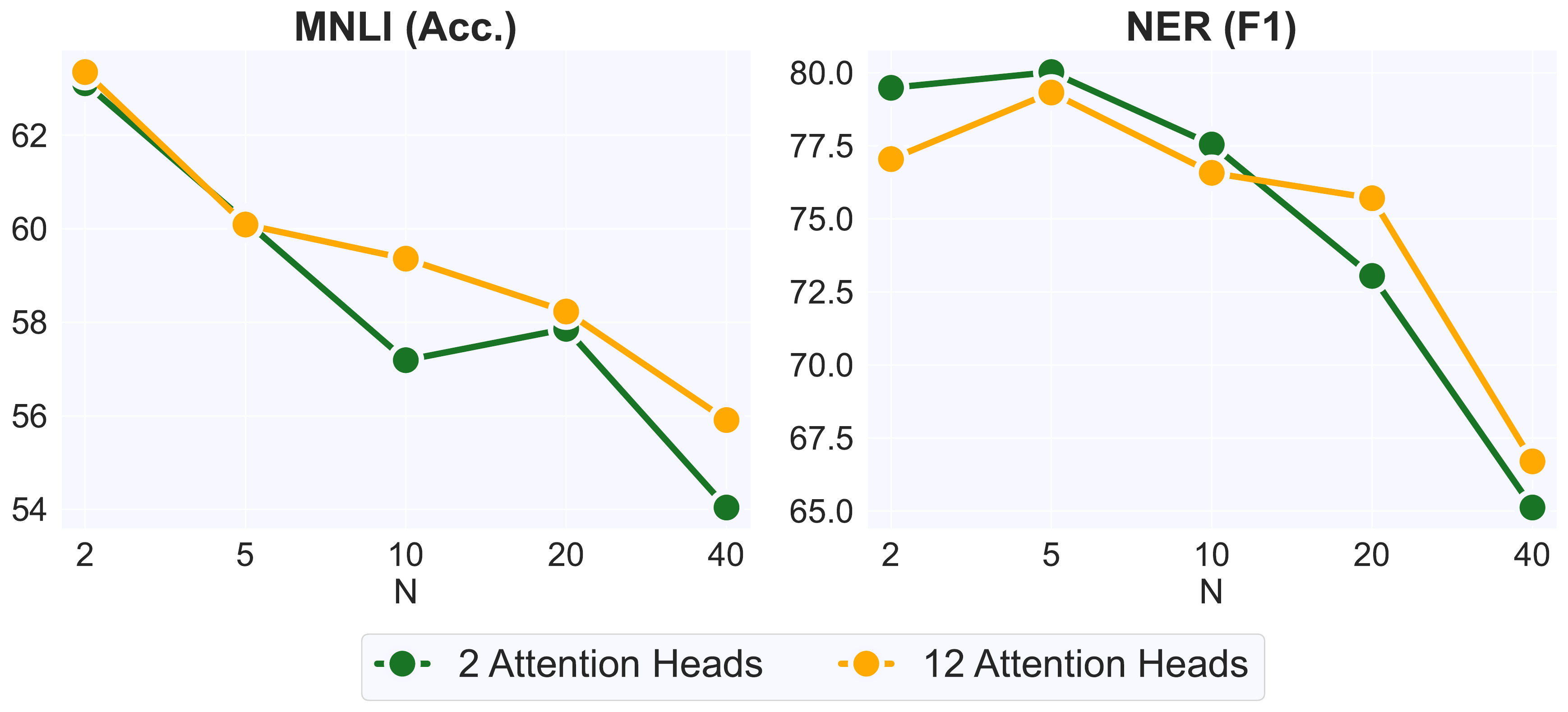}
    \phantomsubcaption
    % \caption{Understanding the role of attention heads in multiplexing. Reducing attention heads to $2$ does not impact performance greatly, suggesting that attention heads are not core to multiplexing.}
    \label{fig:attention_head_ablation}
  \end{subfigure}
  \begin{subfigure}[ht]{0.48\linewidth}
    \includegraphics[width=0.95\linewidth]{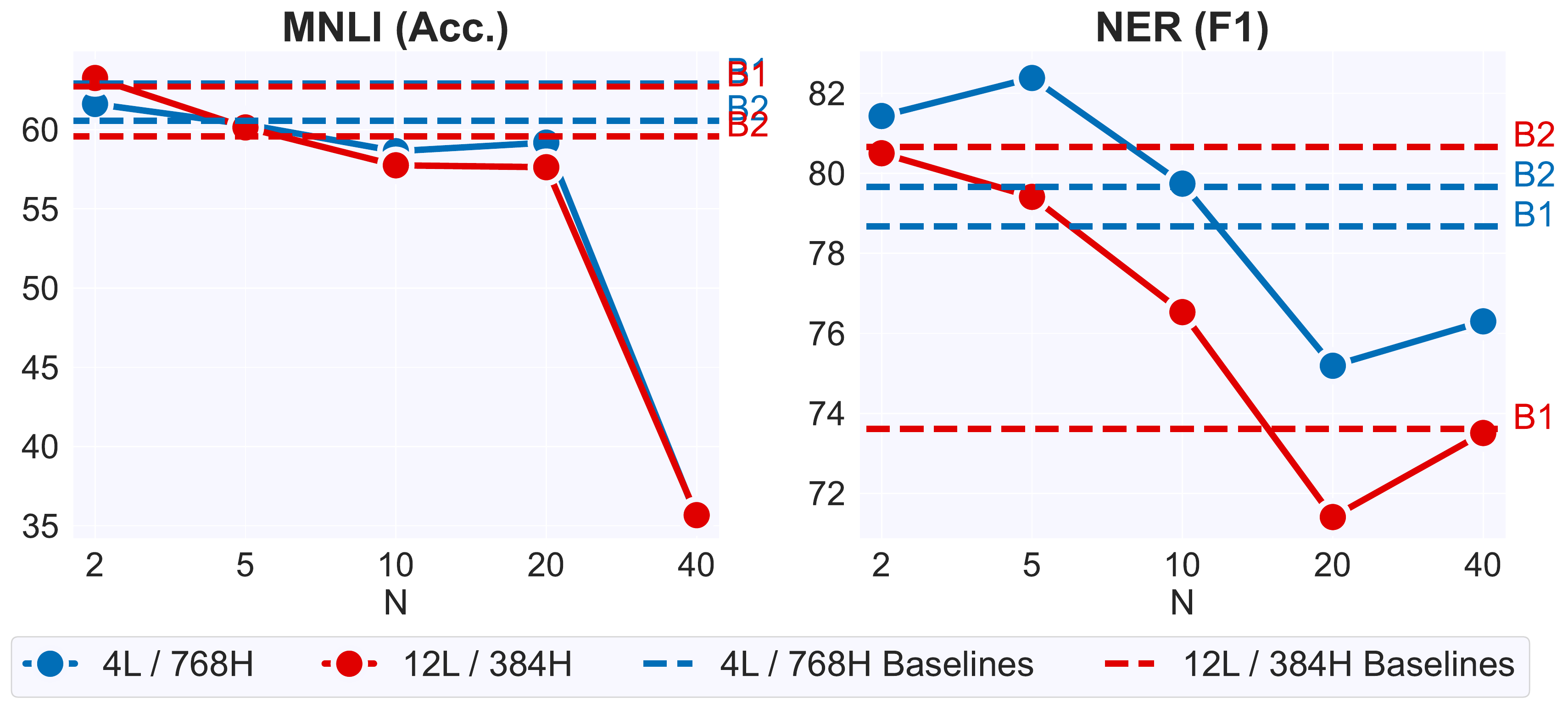}
    \phantomsubcaption
    % \caption{Multiplexing performance with smaller model capacities. Smaller models can multiplex up to $20$ instances without significant drop in performance.}
    \label{fig:smaller_model_mux_chart}
  \end{subfigure}
  \caption{\textbf{(Left a)} Understanding the role of attention heads in multiplexing. Reducing attention heads to $2$ does not impact performance greatly, suggesting that attention heads are not core to multiplexing. \textbf{(Right b)} Multiplexing performance with smaller model capacities. Smaller models can multiplex up to $20$ instances without significant drop in performance.}
\end{figure}

\paragraph{(A1) The number of attention heads seems invariant to multiplexing}
\label{sec:attn_heads}
To understand the role of the number of attention heads in multiplexing, we train a variant of \tmux{} with $2$ self-attention heads per layer. We use \tmux{} with the (Hadamard + Index Embed.) configuration. We find that reducing self-attention heads has minimal effect on performance on the retrieval warm-up task and achieves a retrieval accuracy of $\sim100\%$ up to $N=20$. We then find that on token and sentence-level classification tasks, \tmux{} with 2 self-attention heads performs comparably to \tmux{} with 12 self-attention heads (Figure~\ref{fig:attention_head_ablation}).

% \begin{figure}[ht]
%     \centering
%     \includegraphics[width=0.8\columnwidth]{charts/attention_head_ablation.pdf}
%     \caption{Understanding the role of attention heads in multiplexing. Reducing attention heads to $2$ does not impact performance greatly, suggesting that attention heads are not core to multiplexing.}
%     \label{fig:attention_head_ablation}
% \end{figure}

% \begin{figure}[ht]
%     \centering
%     \includegraphics[width=0.8\columnwidth]{charts/small_model_mux_results.pdf}
%     \caption{Multiplexing performance with smaller model capacities. Smaller models can multiplex up to $20$ instances without significant drop in performance.}
%     \label{fig:smaller_model_mux_chart}
% \end{figure}

% \begin{figure}[ht]
%     \centering
%     \includegraphics[width=0.9\columnwidth]{charts/performance_by_index_hadamard_cond.pdf}
%     \caption{Performance results for different demultiplexing output indices. Results are reported for a Transformer model with the Hadamard product multiplexing and Index Embeddings demultiplexing. Variance of performance across different indices with increasing N.}
%     \label{fig:demux_index_results}
% \end{figure}

\paragraph{(A2) \datamux{} also provides throughput boost with smaller Transformers}
\label{sec:small-model-datamux}
We further investigate the applicability of \datamux{} to smaller Transformer models. We choose two smaller Transformers\footnote{We performed an empirical study to determine models that performed best, see~\ref{appendix:smaller_models} for details.} to multiplex: a $12$ layer with hidden size of $384$ (12L / 384H), a $4$ layer with hidden size of $768$ (4L / 768H). Figure~\ref{fig:smaller_model_mux_chart} shows that these smaller \tmux{} models can also multiplex up to $20$ instances with competitive performance. Figure~\ref{fig:efficiency_chart} illustrates the speedup from the smaller models. As the smaller models can only multiplex up to $20$ instances with reasonable performance, we see that multiplexing with $20$ instances provides an even higher throughput of $25$x, compared to only $18$x for the full-sized \tmux{} with $40$ instances. 

\paragraph{(A3) Performance varies more across different indices as $N$ increases}
The prediction of an instance is conditioned on the index of the instance. Figure~\ref{fig:demux_index_results} illustrates performance on MNLI for different indices across different choices of $N$ for the (Hadamard + Index Embed.) configuration. We observe that performance varies more across different indices for large $N$ (For $N=40$, performance varies across $\sim10$ percentage points).

\begin{figure}[ht]
    \centering
    \includegraphics[width=\columnwidth]{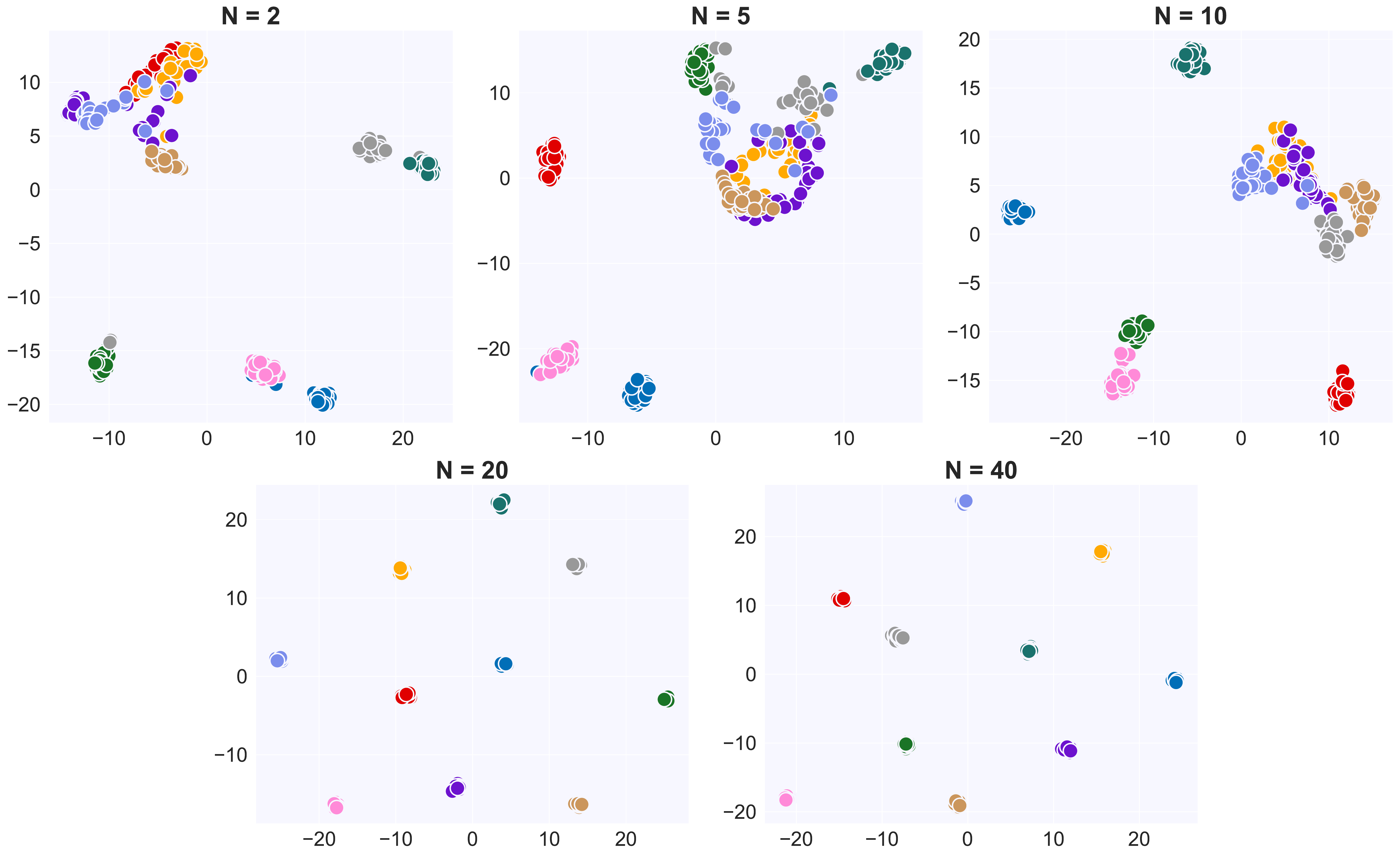}
    \caption{T-SNE plots to understand how the demultiplexed representation of an instance changes with respect to the set of instances it is multiplexed with. Each color (10 in total) a single input sequence. Across different $N$, we find that the demultiplexed representation of an instance is not significantly impacted by the set of instances it is multiplexed with.}
    \label{fig:interference}
\end{figure}

\paragraph{(A4) The demultiplexed representation of an instance is robust to the other instances it is multiplexed with}
To analyze whether the representation of an instance changes with respect to the other samples it is multiplexed with, we randomly select $10$ instances $x_{1}, x_{2}, \cdots x_{10}$ from the MNLI dataset and multiplex each of them (separately) with $30$ different sets of other instances $y_{1}^i, y_{2}^i, \cdots, y_{N-1}^i \text{ for } i \in [1 \cdots 30]$. This generates $30$ different demultiplexed representations for each of the $10$ selected samples. We then visualize the resulting points by reducing the  $768$ dimensional demultiplexed representation to a $200$ dimensional vector with PCA (~\cite{pca}), followed by t-SNE (~\cite{tsne}) visualization. Figure~\ref{fig:interference} shows these clusters for different values of $N$. We find that across different values of $N$, all the points corresponding to the same $x_i$ are very close to each other, which suggests that the representation of an instance is not significantly influenced by the set of instances it is multiplexed with. 

% We also find that the pairwise cosine similarity between vectors in different clusters is close to $0$ for high values of $N$, suggesting that the network is forcing the learned representations of different instances to be orthogonal to each other.

% Training schemes which keep track of these statistics and dynamically scale the losses for different indices might improve performance for large $N$.

\subsection{Theoretical construction for multiplexing in self-attention models}
\label{sec:construction}
To provide some theoretical explanation for why Transformers may be amenable to multiplexing, we also detail below a construction for self-attention based neural networks which  process multiplexed token embeddings in $N$ independent subspaces. Our construction relies on particular structural properties of singular spaces of linear transformations across layers -- we provide a brief sketch below, with more details in Appendix~\ref{sec:app:construction}.

In a model with self-attention, the multi-head attention projects the queries, keys, and values $h$ times with different, learned linear projections to $d_K$, $d_K$ , and $d_V$ dimensions respectively. Thus, given a sequence of $d$-dimensional multiplexed token embeddings $\{\mathbf{w}_t^{1:N}\}_{t\in[1,L]}$, each head looks like:
\begin{equation}
\resizebox{0.45\textwidth}{!}{$\displaystyle
head_i(t) := \sum_{t'=1}^{L}\left[\frac{\exp\left(\frac{\left(\bm W_i^{K}\mathbf{w}^{1:N}_{t'}\right)^\top \bm W_i^{Q} \mathbf{w}^{1:N}_t }{\sqrt{d_K}}\right) }{\sum_{t''}\exp\left(\frac{\left(\bm W_i^{K} \mathbf{w}^{1:N}_{t''}\right)^\top \bm W_i^{Q} \bm u_t }{\sqrt{d_K}}\right)} \bm W_i^{V} \mathbf{w}^{1:N}_{t'}\right] .$}
\end{equation}

Now, 
$\mathbf{w}^{1:N}_{t} = \frac{1}{N}\sum_{k=1}^{N}\phi^{k}(\mathbf{w}^{k}_t)$, and let us assume each function $\phi^{k}$ can be trained to project each embedding into a subspace that is least linearly-dependent with the others.
So, if we define $\mathbf{u}^{(k)}_t := \frac{1}{N}\phi^{k}(\mathbf{w}^{k}_t)$, we assume $\langle\mathbf{u}^{(k)}_t, \mathbf{u}^{(k')}_{t'}\rangle\approx 0$ for all pairs of indices $k\neq k'$ and all positions $t$. To preserve this independent subspace structure after self-attention, we will first need
\begin{equation}
\resizebox{0.45\textwidth}{!}{$\displaystyle
\langle \bm W^V_i \bm u_t^{(k)},  \bm W^V_i \bm u_t^{(k')}  \rangle  = {\bm {u_t^{(k)}}}^{\top} \left({\bm W^V_i}^{\top}\bm W^V_i \right) \bm u_t^{(k')}\approx 0.$}
\label{eq:const-V-condition}
\end{equation}

This is achievable if the eigenvectors of ${\bm W^V_i}^{\top}\bm W^V_i$ can be grouped into $N$ non-overlapping subsets $\{\bm r^{(1)}_{1},\dots, \bm r^{(1)}_{m}\}$, $\cdots,$ $\{\bm r^{(N)}_{1},\dots, \bm r^{(N)}_{m}\}$, where $\bm r_\ell^{(k)}$ are orthonormal vectors (since the Gramian is real symmetric), and span the same input subspaces, denoted as $\mathcal D^1, \dots, \mathcal D^N$. In this case, the vector after transformed by $\bm W^V_i$ can be expressed as a sum of $N$ vectors $\bm v_{i,t}^{(1)},\dots, \bm v_{i,t}^{(N)}$ in dual subspaces $\mathcal D_{V}^1, \dots, \mathcal D_{V}^N$ which are independent of each other. The linear maps from $\mathcal D^k$ to $\mathcal D_V^k$ still allows rich operation on each component of a multiplex input without interference by other components. 

In addition to decompress-able value vectors, we can set the query and key matrices, $\bm W_i^{Q}$ and $\bm W_i^{K}$, to have some subsets of right and left singular vectors such that span $\mathcal D^1, \dots, \mathcal D^N$ and $\mathcal D^1_V, \dots, \mathcal D^N_V$. Then, we can show that the inner product of the query and keys of the $i$-th head can be rewritten as:
\begin{equation}
    \left(\bm W_i^{K}\mathbf{w}^{1:N}_{t'}\right)^\top \bm W_i^{Q} \mathbf{w}^{1:N}_t = \sum_{k=1}^N\tau_{i,t,t'}^{(k)},
\end{equation}
where $\tau_{i,t,t'}^{(k)}$ is a scalar only depending on the $k$-th input sequence. Thus, the self-attention operation at each position can be seen as retrieving values based on the average of query-key similarity scores of $N$ sequences.
\begin{equation}
\resizebox{0.43\textwidth}{!}{$\displaystyle
head_i(t) := \sum_{t'}\left[\frac{\exp\left(\sum_k\tau_{i,t,t'}^{(k)}/\sqrt{d_K}\right) }{\sum_{t''}\exp\left(\sum_k\tau_{i,t,t''}^{(k)}/\sqrt{d_K}\right)} \sum_k \bm v_{i,t}^{(k)}\right]$}
\label{eq:head-construction}
\end{equation}
This average retrieval using soft-max could be a desired property as implicit regularization. However, if we want perfect non-interference in retrieval, the network always has an option to specialize each head to only focus on one input sequence, by setting $\tau^{(k')}_{i,t,t'} = 0$ for all $k'\neq k$, which is easily achievable by controlling singular values of $\bm W_i^{Q}$ or $\bm W_i^{K}$. More details of this construction are provided in Appendix~\ref{sec:app:construction}. 
\vmn{Figure~\ref{sec:attn_heads} suggests that number of attention heads do not significantly impact our empirical multiplexing results. Therefore our theoretical construction does not fully explain what models learn empirically but nonetheless provides an existence proof for multiplexing under certain assumptions.}

\section{Multiplexing for MLPs and CNNs}

We also investigate multiplexing for multilayer perceptrons (MLPs) and convolutional neural networks (CNNs). 
Certain cases of data multiplexing have been explored for convolutional architectures on image classification as techniques for robustness and data augmentation \cite{Rame21, Havasi21, Soflaei20}. These works implicitly employed the frontend layers of a convolutional net as multiplexing layers and suggest that a convolutional neural network can learn at most 3-4 independent subnetworks concurrently \cite{Havasi21}. Since Transformers can be multiplexed for  up to $40$ instances without a severe performance drop, we explore the \datamux{} scheme for MLPs and CNNs on the MNIST image classification task~\cite{Mnist}.\footnote{Implementation details can be found in Appendix~\ref{apendix:cnn_mlp_expt_design}} While our results for these architectures are not as strongly positive as for \tmux{}, we believe that future work on better multiplexing strategies can make them more viable.

% \subsection{Experimental Setup}
% Each image is cropped as $20\times20$ pixels at the center and trained with standard stochastic gradient decent. We also do not apply weight decay or other regularization as these are orthogonal to the multiplexing setting \footnote{Other experimental details can be found in~\ref{apendix:cnn_mlp_expt_design}}. 

% \captionsetup{aboveskip=-10pt}
\begin{figure}[ht]
  \centering
  \begin{subfigure}[ht]{0.54\linewidth}
    \includegraphics[width=0.97\linewidth]{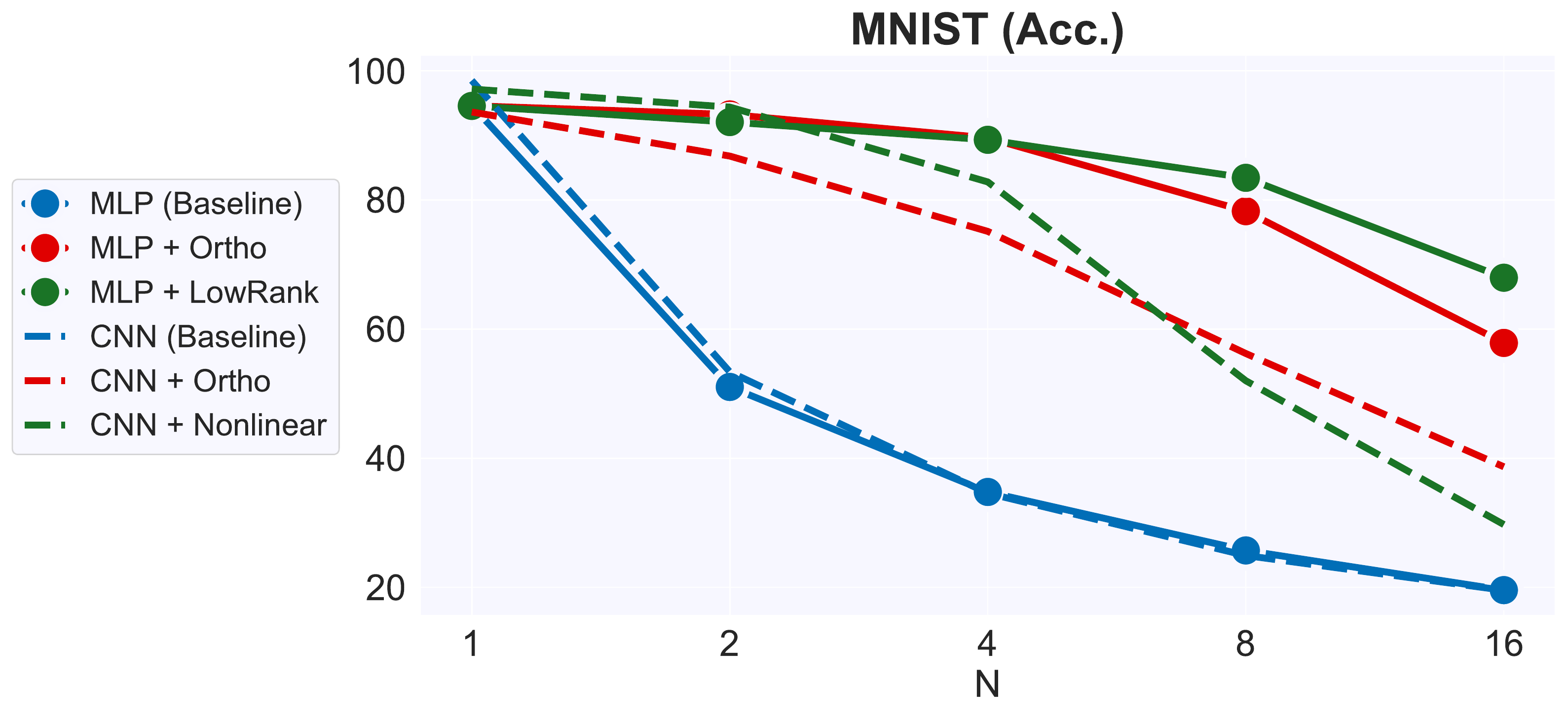}
    \phantomsubcaption
    % \caption{Average test accuracy for multiplexing MLP and CNN for $N=\{1, 2, 4, 8, 16\}$ on the MNIST classification task for different multiplexing strategies. MLPs can be multiplexed up to $N=8$ without significant drop in performance while CNNs can only be multiplexed up to $4$ instances. \kn{maybe say (baseline) for ID models in legend}}
    \label{fig:mlp-vs-cnn}
    \end{subfigure}
%   \hspace{4pt}
%   \begin{subfigure}[ht]{0.23\linewidth}
%     \includegraphics[width=\linewidth]{charts/retrieval_accuracy.pdf}
%     \phantomsubcaption
%     % \caption{Average test accuracy for multiplexing MLP and CNN for $N=\{1, 2, 4, 8, 16\}$ on the MNIST classification task for different multiplexing strategies. MLPs can be multiplexed up to $N=8$ without significant drop in performance while CNNs can only be multiplexed up to $4$ instances. \kn{maybe say (baseline) for ID models in legend}}
%     \label{fig:retrieval_chart}
%     \end{subfigure}
%   \hspace{5pt}
  \begin{subfigure}[ht]{0.45\linewidth}
    \includegraphics[width=0.97\linewidth]{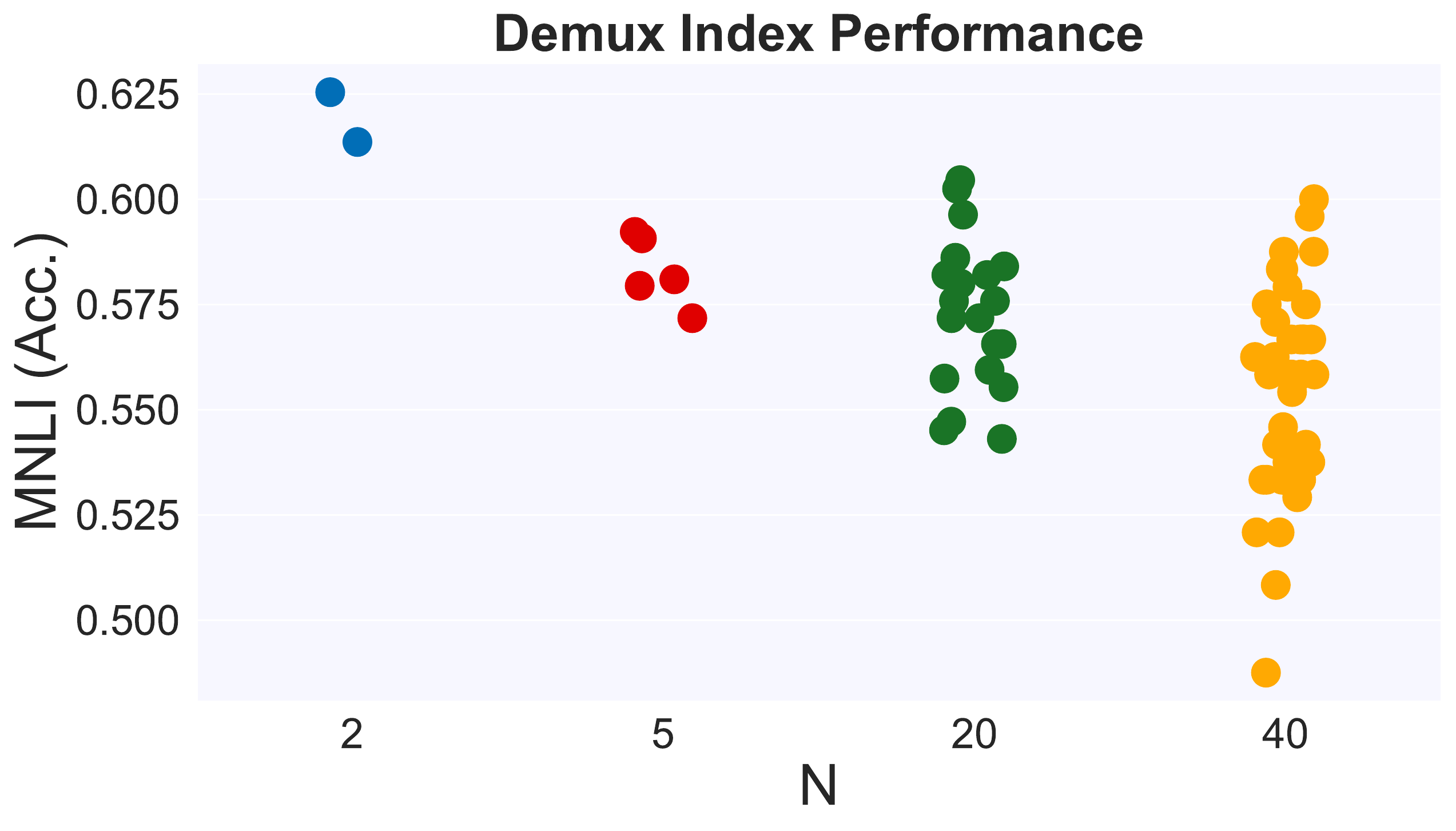}
    \phantomsubcaption
    % \caption{Performance results for different demultiplexing output indices. Results are reported for a Transformer model with the Hadamard product multiplexing and Index Embeddings demultiplexing. Variance of performance across different indices with increasing N.}
    \label{fig:demux_index_results}
    \end{subfigure}
    \vspace{10pt}
    \caption{\textbf{(Left a)} Average test accuracy for multiplexing MLP and CNN for $N=\{1, 2, 4, 8, 16\}$ on the MNIST classification task for different multiplexing strategies. MLPs can be multiplexed up to $N=8$ without significant drop in performance while CNNs can only be multiplexed up to $4$ instances. \textbf{(Right b)} Performance results for different demultiplexing output indices. Results are reported for a Transformer model with the Hadamard product multiplexing and Index Embeddings demultiplexing. Variance of performance across different indices with increasing $N$.}
\end{figure}
\captionsetup{aboveskip=4pt}

% \begin{figure}
%     \centering
%     \includegraphics[width=0.6\columnwidth]{charts/MLPvsCNN-Fig.pdf}
%     \caption{Average test accuracy for multiplexing MLP and CNN for $N=\{1, 2, 4, 8, 16\}$ on the MNIST classification task for different multiplexing strategies. MLPs can be multiplexed up to $N=8$ without significant drop in performance while CNNs can only be multiplexed up to $4$ instances.}
%     \label{fig:mlp-vs-cnn}
% \end{figure}

\paragraph{Multiplexing for MLPs}
Figure~\ref{fig:mlp-vs-cnn} (solid lines) illustrates performance of \datamux{} for MLPs with various multiplexing strategies, paired with the MLP  demultiplexing strategy. As a baseline, we show multiplexing using the identity transformation before combining instances (MLP baseline). Since this transformation does not preserve the order of the multiplexed instances, accuracy expectedly decreases on the order of $1/N$. Multiplexing with random orthogonal matrices (MLP + Ortho) works for up to $8$ instances with an accuracy of $\sim78\%$, compared to $\sim95\%$ for the baseline without multiplexing. Since random orthogonal projections cannot separate inputs perfectly, we also experiment with a set of $N$ low-rank independent transformations for multiplexing  (MLP + LowRank) and observe that this improves performance slightly for larger $N$ (by $5 \%$ for $N$=8).

\paragraph{Multiplexing for CNNs}
Results for CNN multiplexing are shown in Figure~\ref{fig:mlp-vs-cnn} (dashed lines). Like in the MLP case, all methods shown use the MLP demultiplexing strategy. We notice that the baselines (CNN baseline) has a roughly $N$-fold performance drop because of the unidentifiability of input order. The ``Ortho'' transformation performs quite poorly for CNNs (only $\sim56\%$ for $N=8$). This is likely because this transformation destroys the property of spatial locality that CNNs rely on. We therefore also explore $N$ two-layer convolutional networks with a \textit{tanh} activation as out multiplexing transformations $\phi^i$  (CNN + Nonlinear). We note that even with this transformation, the dimensionality of the multiplexed representation still remains equal to the dimensionality of a single input. 
We find that for $N\leq 4$, performance is above $80\%$, which is significantly better than CNN+Ortho, but performance drops rapidly for $N > 4$. We show results of other multiplexing strategies for CNN in Appendix \ref{appendix:cnn}.

Overall, multiplexing for MLPs and CNNs seems to be more challenging than Transformers as evidenced by the sharper performance drops with increasing $N$. However, we believe these numbers are still non-trivial and show the potential for multiplexing in CNNs and MLPs, with multiplexing and demultiplexing strategies better suited for these architectures.

\section{Discussion}
\label{sec:discussion_section}
% We introduce \datamux{}, an entirely novel setting for neural network training and inference, and demonstrate its viability using a variety of network architectures and tasks. Our results reveal the surprising ability of neural networks to predict on up to 40 input instances simultaneously using data multiplexing. Our efforts have sought to show the potential for data multiplexing using methods that require limited computational overhead or added learned parameters; achieving results that incur minimal degradation in performance while simultaneously increasing system throughput dramatically. While we've focused here on showing the ability of neural networks to use \datamux{} generally, we believe there are many other potential applications for \datamux{}, especially large-scale pre-training, multi-modal processing, and multi-lingual models.

% Our theoretical analysis of data multiplexing for self-attention networks, in conjunction with our many empirical analyses investigating the capabilities and effects of different design choices for \datamux{} shows that data multiplexing works in a variety of conditions. The precise architectural and data conditions enabling \datamux{}, however, remains unclear. We hope that future work can further elucidate the nature and limits of data multiplexing for neural networks.

In this work, we have shown that neural networks can be trained to multiplex, i.e. process multiple inputs using a single ``mixed'' representation.
We introduced \datamux{}, a novel setting for training and inference with multiplexing, and demonstrate its viability using a variety of network architectures and tasks. Our results reveal the surprising ability of neural networks to predict on up to $40$ input instances simultaneously using data multiplexing. Our efforts have sought to show the potential for data multiplexing using methods that require limited computational overhead or added learned parameters; achieving results that incur minimal degradation in performance while simultaneously increasing system throughput dramatically. 

While the scope of this paper is to demonstrate the general ability and potential for neural networks to use data multiplexing, there are several avenues for future research to investigate the limits and theoretical implications of \datamux{}. Directions of particular interest include: large-scale pre-training, multi-modal processing, multi-lingual models, different mechanisms of multiplexing and demultiplexing, and a more rigorous understanding of the architectural, data, and training conditions enabling data multiplexing.

% We thank, in alphabetical order, Ameet Deshpande, Tianyu Gao, Jens Tuyls, Mengzhou Xia, and Shunyu Yao for their valuable comments and feedback.

% Our theoretical analysis of data multiplexing for self-attention networks, in conjunction with our many empirical analyses investigating the capabilities and effects of different design choices for \datamux{} shows that data multiplexing works in a variety of conditions. The precise architectural and data conditions enabling \datamux{}, however, remains unclear. We hope that future work can further elucidate the nature and limits of data multiplexing for neural networks.

% conclusion
    % muxing is possible upto 40 instances with minimal loss in accuracy with almost 20x speedup
% note on pretraining: show that we can mux broadly across acrchitectures and model sizes, Pretrain and finetune paradigm,
% 
% future work ideas
    %
    %s
    %
    %
    %

\section{Acknowledgements}
We thank Ameet Deshpande, Shunyu Yao, Jens Tuyls, Tianyu Gao and Mengzhou Xia for their valuable feedback on early drafts and encouragement throughout the course of this project, and the anonymous reviewers for their suggestions on improving the paper.

\bibliography{references.bib}
%%%%%%%%%%%%%%%%%%%%%%%%%%%%%%%%%%%%%%%%%%%%%%%%%%%%%%%%%%%%
\section*{Checklist}

\begin{enumerate}

\item For all authors...
\begin{enumerate}
  \item Do the main claims made in the abstract and introduction accurately reflect the paper's contributions and scope?
    \answerYes{}
  \item Did you describe the limitations of your work?
    \answerYes{Refer to~\ref{appendix:limitations}}
  \item Did you discuss any potential negative societal impacts of your work?
    \answerYes{Refer to~\ref{appendix:negative_impact}}
  \item Have you read the ethics review guidelines and ensured that your paper conforms to them?
    \answerYes{}
\end{enumerate}

\item If you are including theoretical results...
\begin{enumerate}
  \item Did you state the full set of assumptions of all theoretical results?
    \answerYes{}
        \item Did you include complete proofs of all theoretical results?
    \answerYes{}
\end{enumerate}

\item If you ran experiments...
\begin{enumerate}
  \item Did you include the code, data, and instructions needed to reproduce the main experimental results (either in the supplemental material or as a URL)?
    \answerYes{Refer to supplementary material}
  \item Did you specify all the training details (e.g., data splits, hyper-parameters, how they were chosen)?
    \answerYes{Refer to~\ref{appendix:transformer_expt_details} and ~\ref{apendix:cnn_mlp_expt_design}}
        \item Did you report error bars (e.g., with respect to the random seed after running experiments multiple times)?
    \answerYes{Refer to~\ref{appendix:variance}}
        \item Did you include the total amount of compute and the type of resources used (e.g., type of GPUs, internal cluster, or cloud provider)?
    \answerYes{Refer to~\ref{appendix:throughput_expt_details}}
\end{enumerate}

\item If you are using existing assets (e.g., code, data, models) or curating/releasing new assets...
\begin{enumerate}
  \item If your work uses existing assets, did you cite the creators?
    \answerYes{}
  \item Did you mention the license of the assets?
    \answerNA{}
  \item Did you include any new assets either in the supplemental material or as a URL?
    \answerNA{}
  \item Did you discuss whether and how consent was obtained from people whose data you're using/curating?
    \answerNA{}
  \item Did you discuss whether the data you are using/curating contains personally identifiable information or offensive content?
    \answerNA{}
\end{enumerate}

\item If you used crowdsourcing or conducted research with human subjects...
\begin{enumerate}
  \item Did you include the full text of instructions given to participants and screenshots, if applicable?
    \answerNA{}{}
  \item Did you describe any potential participant risks, with links to Institutional Review Board (IRB) approvals, if applicable?
    \answerNA{}
  \item Did you include the estimated hourly wage paid to participants and the total amount spent on participant compensation?
    \answerNA{}
\end{enumerate}

\end{enumerate}
\clearpage
\pagenumbering{arabic}% resets `page` counter to 1
\renewcommand*{\thepage}{A\arabic{page}}
\appendix
\section{Appendix}
\label{sec:appendix}

\subsection{Ethical considerations}
\vmn{\datamux{} has the potential to provide great computational efficiency when used in real production systems, where queries across different users can be aggregated to get simultaneous predictions for different users. However, this also present potential risks and scope for misuse. First, the multiplexing layer might `leak' data between different users, which could potentially lead to privacy concerns. Second, the output of an instance might be influenced by other instances in the multiplexing batch, which may present the possibility of black-box attacks to manipulate information, especially in the multi-user setting. Finally, \datamux{} models predictions could be harder to interpret with current techniques since the model's internal representations depend on the set of instances it was multiplexed with.
}

\vmn{We believe these are important and interesting research problems to solve technically and may also require careful policy development for deploying these models (e.g. restricting multiplex batches to single users).}

\subsection{Limitations}
Here, we list some of the limitations of this work:
\begin{itemize}
    \item \textbf{Convergence time increases with increasing number of instances:} Training multiplexed models with large number of instances takes larger number of iterations for convergence as increasing number of instances increases the difficulty of multiplexing and demultiplexing. We hope that future work can explore different multiplexing and demultiplexing strategies to improve rate of convergence during training.
    \item \textbf{Multiplexing on CNNs and MLPs not as strong as that of the Transformer:} While we demonstrate non-trivial multiplexing capabilities for the CNN and MLP architectures with our approach, the results are not as strong as that of the Transformer architecture. We hope future work will develop better multiplexing approaches for these architectures.
\end{itemize}
\label{appendix:limitations}

% \vmn{This does pose ethical concerns as the output of an instance might be influenced by which instances it was multiplexed with and therefore would be less faithful to the input instance. This could also compound the effect of spurious correlations as these spurious correlations might "leak" between different instances and make it harder for the model to generalize to unseen samples. Moreover, this could also make the model's predictions much less interpretable as the model's output could be partially conditioned on the set of instances it was multiplexed with. } 
\label{appendix:negative_impact}

\subsection{Theoretical construction of multiplexing transformer}
\label{sec:app:construction}
Following the same notations defined in Section \ref{sec:construction},
assume that the components of a multiplexed input, $\mathbf{u}_t^{(1)},$ $\dots,$ $\mathbf{u}_t^{(N)}$, all approximately lie in $N$ linearly independent subspaces $\mathcal D^1$, $\dots$,$\mathcal D^N$. Equation \ref{eq:const-V-condition} is realizable when eigenvectors of ${\bm W^V_i}^{\top}\bm W^V_i$ can be grouped into $N$ non-overlapping subsets $\{\bm r^{(1)}_{1},\dots, \bm r^{(1)}_{m}\}$, $\cdots,$ $\{\bm r^{(N)}_{1},\dots, \bm r^{(N)}_{m}\}$, where $\bm r_\ell^{(k)}$ are orthonormal vectors (since the Gramian is real symmetric), and span the same input subspaces. In this case, the vector after linear transformation $\bm W^V_i$ can be expressed as a superposition of $N$ vectors in $N$ independent subspaces $\mathcal D_{V}^1, \dots, \mathcal D_{V}^N$. We now verify this statement.

\textit{Proof.} Since eigenvectors of ${\bm W^V_i}^{\top}\bm W^V_i$ span the same subspaces, we can write each component of $\mathbf{w}^{1:N}_t$ as a linear combination of the corresponding subset of eigenvectors,
\begin{equation}
\mathbf{w}^{1:N}_t= \sum_{k}\mathbf{u}_t^{(k)} = \sum_{k,\ell} \alpha^{(k)}_{\ell,t} \bm r^{(k)}_{\ell}.
\end{equation}
Let $\bm W^V_i = \bm L\bm \Sigma\bm R^{\top}$be the singular decomposition of $\bm W^V_i$, where $\bm L \in \mathbb R^{d_V\times d_V}$, $\bm R \in \mathbb R^{d \times d}$ are two orthogonal matrices, and $\bm\Sigma$ is a $d_V \times d$ rectangular diagonal matrix with non-negative real numbers on the diagonal. For each column of $\bm R$ that contributes to input subspaces, e.g., $\bm r_{\ell}^{(k)}$, we denote the dual left singular vector at the corresponding column of $\bm L$ as $\bm l_{\ell}^{(k)}$, and the singular value at the corresponding column and row $\sigma_\ell^{(k)}$
\begin{eqnarray}
\bm v_{i,t} & = & \bm W^V_i \mathbf{w}^{1:N}_t = \bm L\bm \Sigma\bm R^{\top}\sum_{k,\ell}\alpha^{(k)}_{\ell,t}\bm r_\ell^{(k)} \\
& = & \sum_{k=1}^{N}\left(\sum_{\ell} \alpha_{\ell,t}^{(k)}\sigma_{\ell}^{(k)}\bm l_{\ell}^{(k)} \right).
\end{eqnarray}
Since $\bm l_{\ell}^{(k)}$ are columns of $\bm L$ that are orthogonal to each other, $\bm v_{i,t}$ above is a superposition of $N$ vectors $\bm v_{i,t}^{(k)}$ in independent subspaces, and each subspace is defined by
\begin{equation}
\mathcal D^{V}_k \cong {\tt span}\{\bm l_{\ell}^{(k)}\}.
\end{equation}
$\blacksquare$

In addition to decompress-able value vectors, we can set the linear transformations for queries and keys to have the same singular space structures to prevent complicated interference. More specifically, assume that $\bm W_i^{Q}$ and $\bm W_i^{K}$ have some subsets of right and left singular vectors such that
\begin{equation}
\mathcal D_{k} \cong {\tt span} \{ {\bm r^{Q}_\ell}^{(k)}\} \cong {\tt span} \{{\bm r^{K}_\ell}^{(k)}\},
\end{equation}
and
\begin{equation}
\mathcal D^K_{k} \cong {\tt span} \{ {\bm l^{Q}_\ell}^{(k)}\} \cong {\tt span} \{ {\bm l^{K}_\ell}^{(k)}\}.
\end{equation}
The inner product between the query and keys can be rewritten as 
\begin{align}
&\left(\bm W_i^{K}\mathbf{w}^{1:N}_{t'}\right)^\top \bm W_i^{Q} \mathbf{w}^{1:N}_t\\ 
&= \resizebox{0.35\textwidth}{!}{$\displaystyle\left(\sum_{k,\ell} \beta_{\ell,t'}^{(k)}{\sigma^K_{\ell}}^{(k)}{\bm l_{\ell}^K}^{(k)} \right)^{\top}\left(\sum_{k,\ell} \gamma_{\ell,t}^{(k)}{\sigma^Q_{\ell}}^{(k)}{\bm l_{\ell}^Q}^{(k)} \right)$}\\
&= \resizebox{0.35\textwidth}{!}{$\displaystyle\sum_{k=1}^{N}\left[\sum_{\ell,\ell'} \beta_{\ell',t'}^{(k)}\gamma_{\ell,t}^{(k)}{\sigma^K_{\ell'}}^{(k)}{\sigma^Q_{\ell}}^{(k)}\left({\bm l_{\ell'}^K}^{(k)}\right)^{\top}{\bm l_{\ell}^Q}^{(k)}\right]$}\\
&=\sum_{k=1}^N\tau_{t,t'}^{(k)}
\end{align}
where $\tau_{t,t'}^{(k)}$ is a scalar only depending on the $k$-th input sequence (for simplification we omit head index $i$).
Thus, the self-attention operation at each position can be seen as retrieving values based on the average of query-key similarity scores of $N$ sequences.
\begin{equation}
\resizebox{0.42\textwidth}{!}{$\displaystyle
head_i(t) := \sum_{t'}\left[\frac{\exp\left(\sum_k\tau_{i,t,t'}^{(k)}/\sqrt{d_K}\right) }{\sum_{t''}\exp\left(\sum_k\tau_{i,t,t''}^{(k)}/\sqrt{d_K}\right)} \sum_k \bm v_{i,t}^{(k)}\right]$} .\end{equation}
The average retrieval by soft-max could be undesirable. However, this will not affect the quality of decompression at all. If we want perfect non-interference in retrieval, one always has an option to specialize each head to only focus on one input sequence, by setting $\tau^{(k')}_{i,t,t'} = 0$ for all $k'\neq k$. This is easily achievable by controlling singular values of $\bm W_i^{Q}$ or $\bm W_i^{K}$.

Finally, we concatenate all heads and linearly project the output with $\bm W^O \in \mathbb R^{d\times hd_V}$ to a $d$-dimensional space again. This step is exactly equivalent to having $h$ projection matrices $\bm W^O_i \in \mathbb R^{d\times d_V}$ acting on different heads and aggregating resulting vectors.
\begin{equation}
\resizebox{0.42\textwidth}{!}{$\displaystyle
output(t) = \bm W^O \left(\ \|_{i=1}^{h}head_i(t)\right) = \sum_{i=1}^{h}\bm W^O_i head_i(t)
$}  
\end{equation}
Employing a similar structure for right singular vectors as $\bm W^V_i$, we can make each $\bm W^O_i$ also preserves the independence of subspaces.

\subsection{Multiplexing with Hadamard + Ortho strategy trains consistently for different runs}
\label{appendix:variance}
\begin{figure}[ht]
\begin{subfigure}[ht]{0.48\linewidth}
    \centering
    \includegraphics[width=\columnwidth]{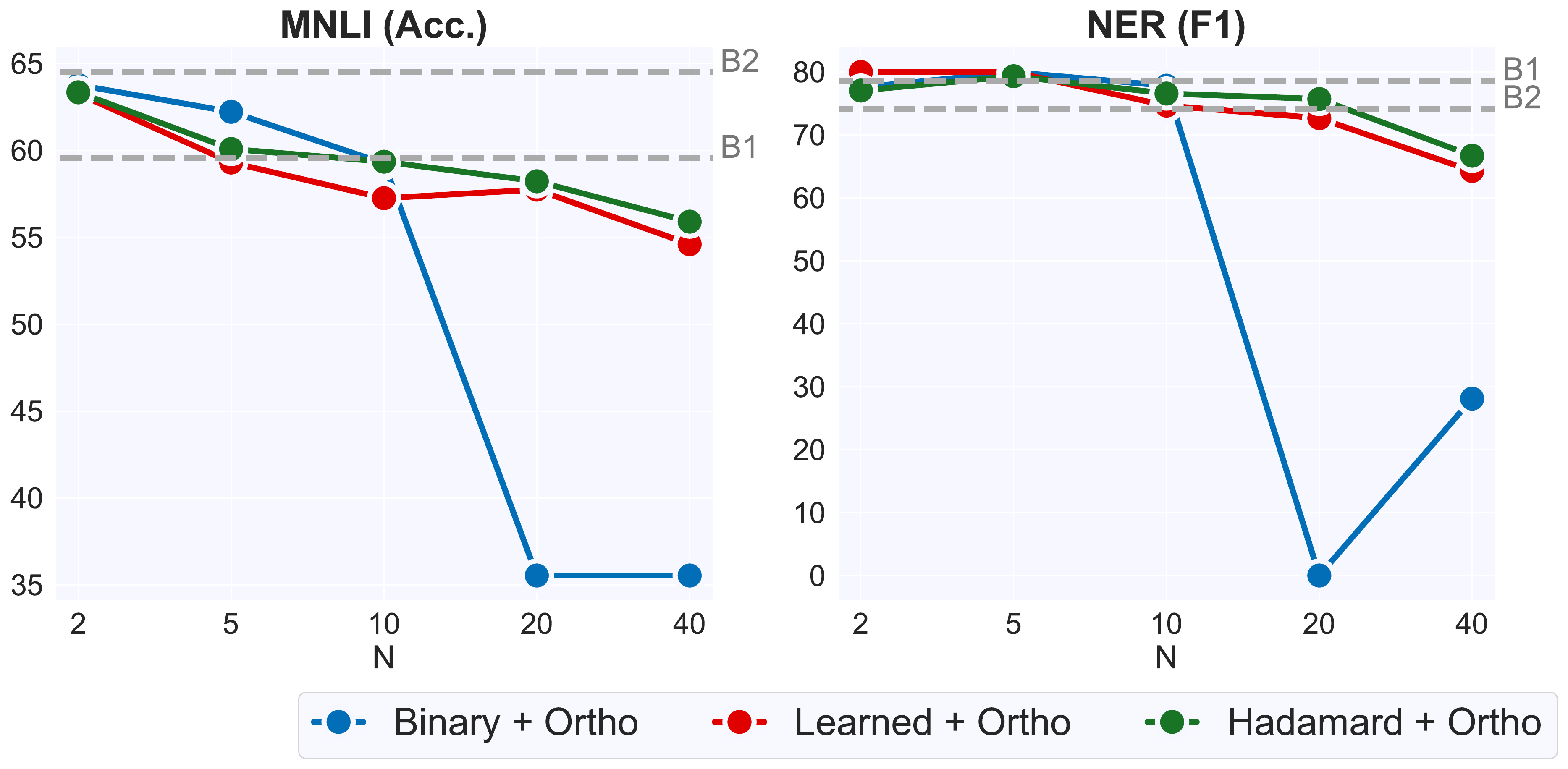}
    \phantomcaption
    \label{fig:superposition_ablation_chart}
\end{subfigure}
\hspace{6pt}
\begin{subfigure}[ht]{0.48\linewidth}
    \centering
    \includegraphics[width=\columnwidth]{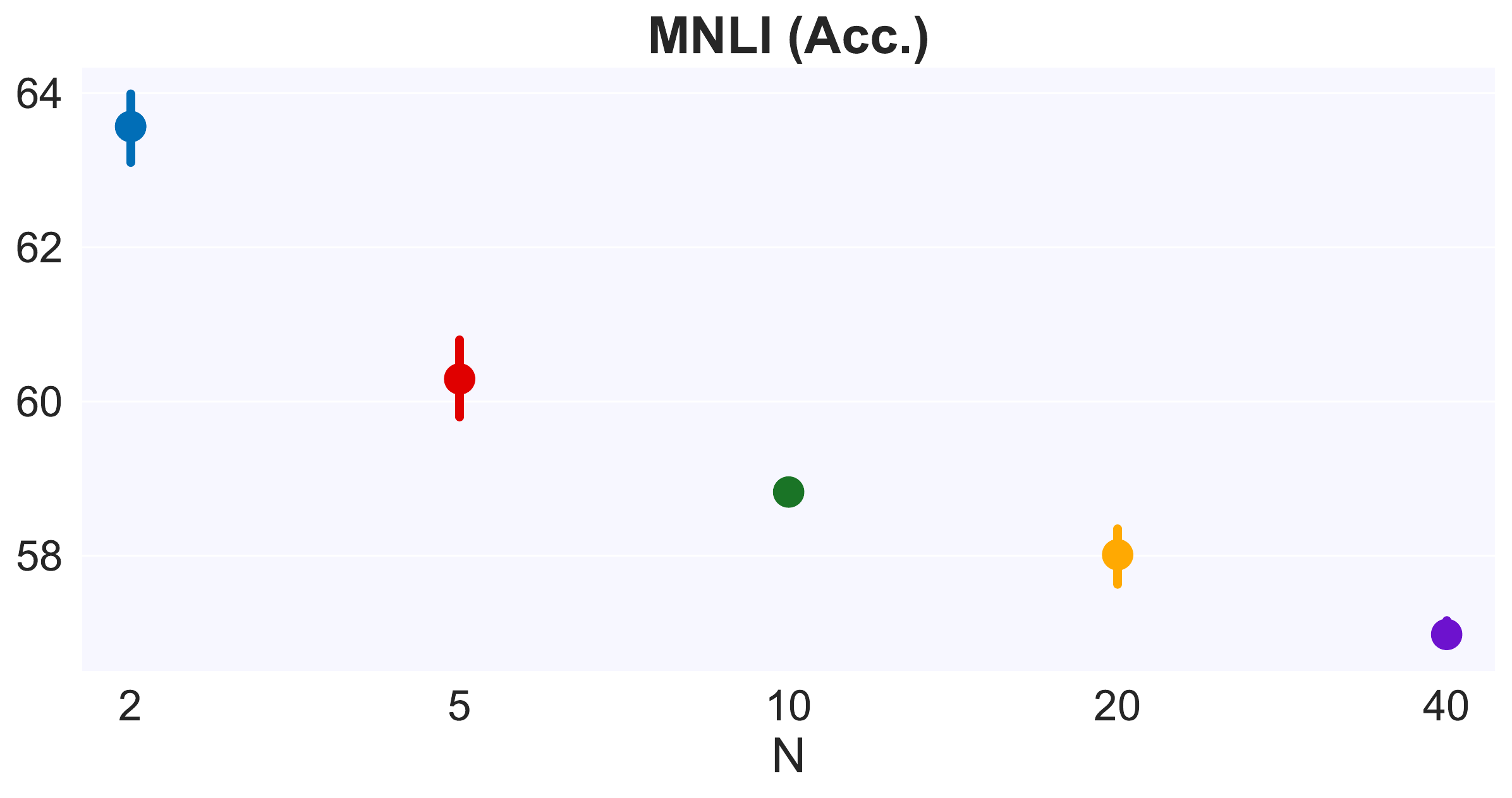}
    \phantomcaption
    \label{fig:mnli_variance}
\end{subfigure}
\caption{\textbf{(Left a)} Evaluation on alternative superposition methods for multiplexing. Binary vectors fail to multiplex beyond $10$ instances, while unfreezing Gaussian vectors for multiplexing does not help. \textbf{(Right b)} MNLI accuracy results with error bar for the Hadamard + Ortho strategy across 3 different random seeds. Variance in performance is minimal to none.}
\end{figure}

In Figure~\ref{fig:mnli_variance} we show MNLI performance for the Hadamard + Ortho multiplexing strategy with error bars across 3 random seeds. We observe that the value of $N$ has no effect on variance in performance across seeds, with all values showing minimal variance. Since we use the same retrieval pre-trained weights, the random seed only affects the demultiplexer and classification head initializations.

\subsection{Alternate multiplexing strategies}
\label{appendix:superpose_ablations}
We experiment with different multiplexing strategies and use index embeddings for demultiplexing. For the Hadamard product, we try unfreezing the random Gaussian vectors and update through optimization (``Learned''). We also experiment with binary masking, where the $i^{th}$ binary vector selects the $i^{th}$ chunk of size $d / N$ from the input representation of instance $i$ (``Binary''). We first look at the performance on the retrieval warm-up task. Figure~\ref{fig:retrieval_chart} shows that the unfreezing the vectors does not significantly change performance. We also observe that binary vectors fails to multiplex for large $N$, suggesting the multiplexing is more capable than just concatenating multiple downsampled inputs of $d / N$ dimension when $N$ is large. We see similar trends for the MNLI and NER (illustrated in Figure~\ref{fig:superposition_ablation_chart}) with unfrozen vectors not impacting performance and binary vectors failing to multiplex for large $N$.

\begin{figure*}[ht]
    \centering
    \includegraphics[width=\textwidth]{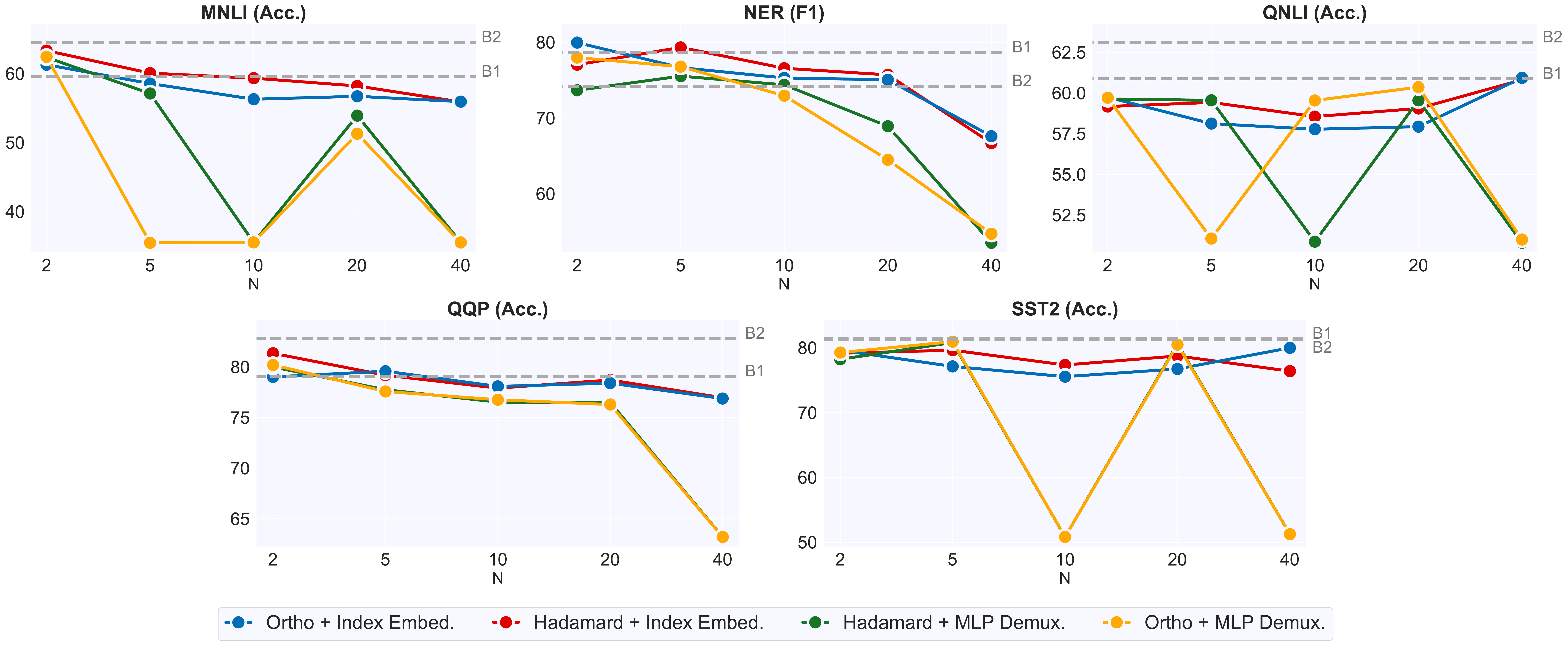}
    \caption{Primary results for NLP tasks mirroring those shown in Figure \ref{fig:large_results_chart}. We additionally show results for the MLP Demux. strategy. While MLP Demux. demultiplexing very works well for retrieval, shown in Figure \ref{fig:retrieval_chart}, we find that it typically performs slightly worse than the Index Embedding method and it leads to unstable optimization. This method also leads to an increase in parameter size proportional to $N$, since we must now train $N$ independent MLPs for each input index.}
    \label{fig:appendix_large_results_chart}
\end{figure*}

\subsection{MLP Demux Results}
\label{appendix:mlp_demux}
We show our primary results for NLP tasks in Figure \ref{fig:large_results_chart}, though we've excluded models using the MLP Demux. demultiplexing method. We provide these results instead in Figure \ref{fig:appendix_large_results_chart} to highlight the optimization instability encountered during training of models using the MLP Demux. method. Especially curious is the failure to converge at apparently arbitrary points for $N$, such as converging for $N=20$ for MNLI yet not converging for $N=10$ despite $N=10$ being a presumably simpler setting.

\subsection{Smaller models achieve good performance across tasks}
\label{appendix:smaller_models}
\begin{figure}[ht]
    \centering
    \begin{subfigure}[ht]{0.48\linewidth}
    \includegraphics[width=\columnwidth]{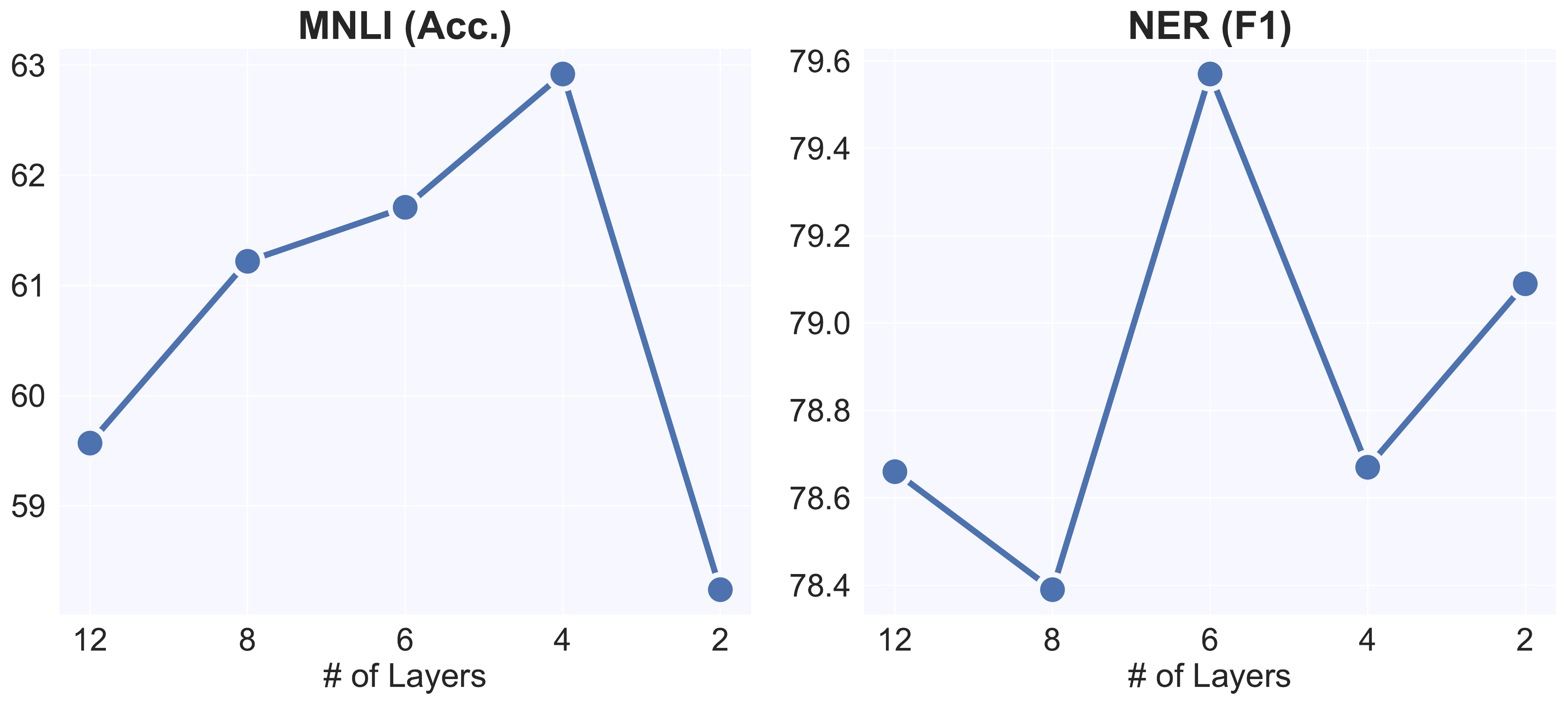}
    \end{subfigure}
    \begin{subfigure}[ht]{0.48\linewidth}
    \includegraphics[width=\columnwidth]{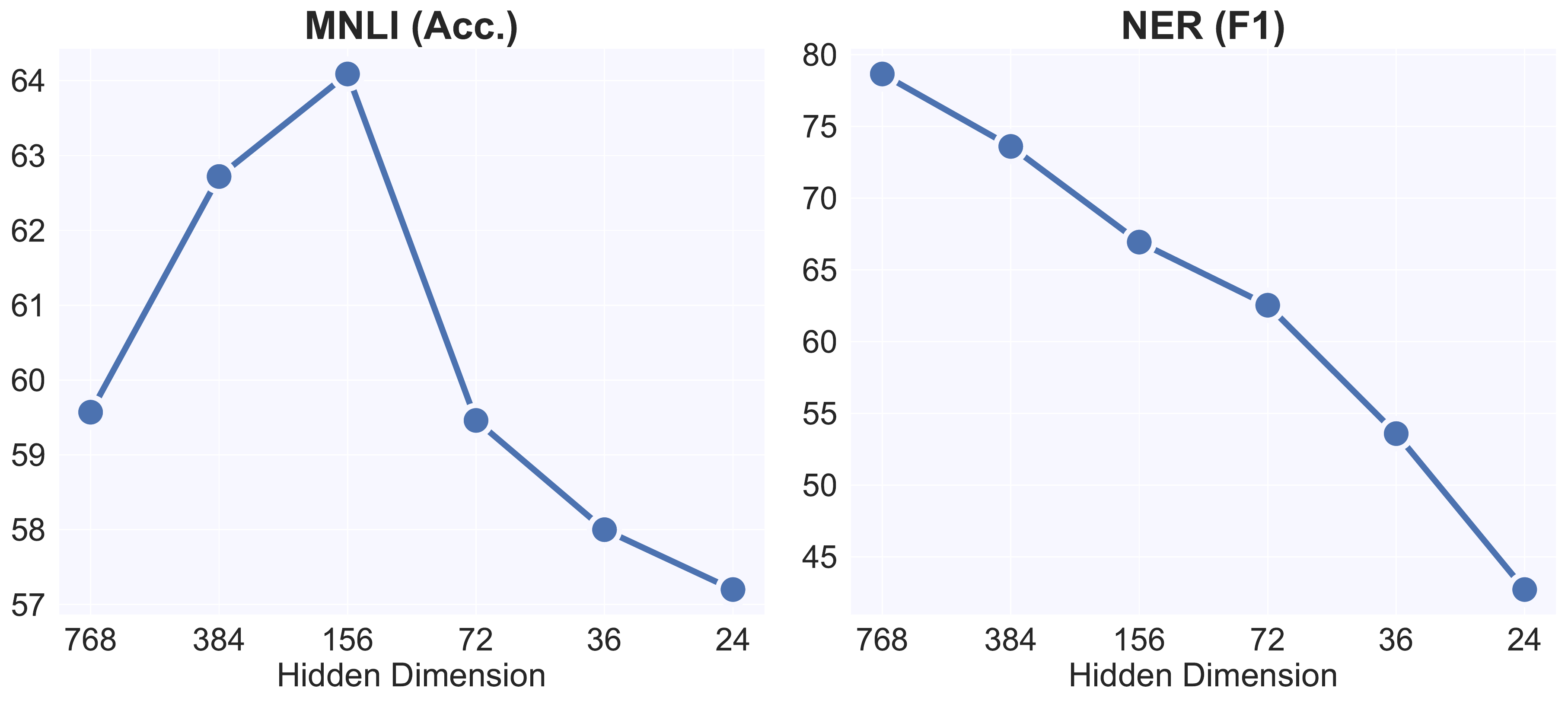}
    \end{subfigure}
    \caption{Model performance for different hidden dimension sizes and number of layers.}
    \label{fig:ablation_charts}
\end{figure}

For our evaluation tasks, our base model might be over-parametrized and smaller models might perform equally well. Figures~\ref{fig:ablation_charts} shows performance on MNLI and NER as we vary the hidden size and the number of layers in the transformer. We observe that smaller models are competitive on both tasks and in the following section explore the possibility of getting higher throughput by multiplexing smaller models. 
\subsection{Experiment details for computing throughput}
We compute throughput over $20000$ samples on the MNLI task on a single Nvidia RTX 2080 machine. We use $4$ batch sizes and take the maximum value.  
\label{appendix:throughput_expt_details}
\subsection{Implementation details for multiplexing Transformers}
We train all models to convergence. We use a learning rate of $2e-5$ and $5e-5$ for baselines and report the best performance. For multiplexed models, we use a learning rate of $5e-5$. However, for large $N$, we use $2e-5$ in case learning rate of $5e-5$ does not converge. We use a batch size of $32$ for the baselines and use slightly smaller batch sizes for multiplexing as multiplexing effectively increase our size of the batch and therefore we need to keep more input instances in memory, leading to a drop in batch size. For the language tasks, we report numbers on the validation split and do not perform any extensive hyper-parameter sweeps.
\label{appendix:transformer_expt_details}
\subsection{Experiment Design details for CNNs and MLPs}
\label{apendix:cnn_mlp_expt_design}
Each image is cropped as $20\times20$ pixels at the center and trained with standard stochastic gradient decent. We also do not apply weight decay or other regularization as these are orthogonal to the multiplexing setting.

Principle component analysis on all $60000$ training images indicates that $86.54\%$ variance can be explained by top $50$ principle components, suggesting that if we only keep the top $50$ dimensions for each input image and project them into linearly independent subspaces, ideally we can multiplex $8 (d= 400 / 50)$ inputs as just one input without much information loss. We test the model performances with $N = \{1, 2, 4, 8, 16\}$ multiplexed inputs.

Our MLP consists of a hidden layer with 100 neurons, a demultiplexing layer that maps the 100 hidden neurons to $20\times N$ neurons, and a shared linear readout layer that maps each group of 20 neurons to 10 categories for classification. Our CNN is similar to the classic LeNet \cite{lecun90}, consisting of three convolutional layers with 10 activation maps from 3x3 kernels, 16 activation maps from 4x4 kernels and 120 activation maps from 3x3 kernels, one linear layer maps to 84 hidden neurons, a demultiplexing layer that maps them to $84\times N$ neurons, and a shared linear readout layer that maps each group of 84 neurons to 10 categories for classification. The first two convolutional layers are followed by 2x2 max pooling. Every linear layer or convolution layer in our models is followed by a $\tanh$ activation. Labels are $+1$ for the correct digit and $-1$ for the incorrect digits. And we use the mean squared error (MSE) loss to train all our models.

We train all our models with the standard stochastic gradient decent with fixed learning rates, and the batch size is 32.  

To generate the low-rank approximations (``LowRank'' in Figure \ref{fig:mlp-vs-cnn}), we divide $d$ random orthogonal row vectors into $N$ groups, and multiplying them by another $d\times d$ orthogonal matrix.

\subsection{Other multiplexing strategies for CNNs}
\label{appendix:cnn}

To make the multiplexing method more compatible with CNNs, we first tried some simple separation functions for images. 2D rotations (\text{SO}(2)) work much better than $\text{SO}(d)$ when $N \leq 2$. When $N=1$, rotating inputs certainly does no harm to the performance. When $N=2$, there is some unavoidable interference between two rotated images, while CNN can still easily distinguish inputs with decent accuracy. However, when $N > 2$, inputs are heavily overlapped, and CNN fails to distinguish many digits. The accuracy among different inputs also varies largely, with std. at 10-24\% when $N\geq 8$, indicating the permutation symmetry of inputs is hard to preserve during optimization. We also tried other simple 2D transformations like mosaic and downsampling so that all inputs are perfectly separated but get blurred. CNN can only answer one of the inputs correctly for this mosaic transformation, since we are only testing the vanilla convolutional architecture, which is not suitable for object detection without proposing bounding boxes.

\begin{figure}
    \centering
    \includegraphics[width=0.6\columnwidth]{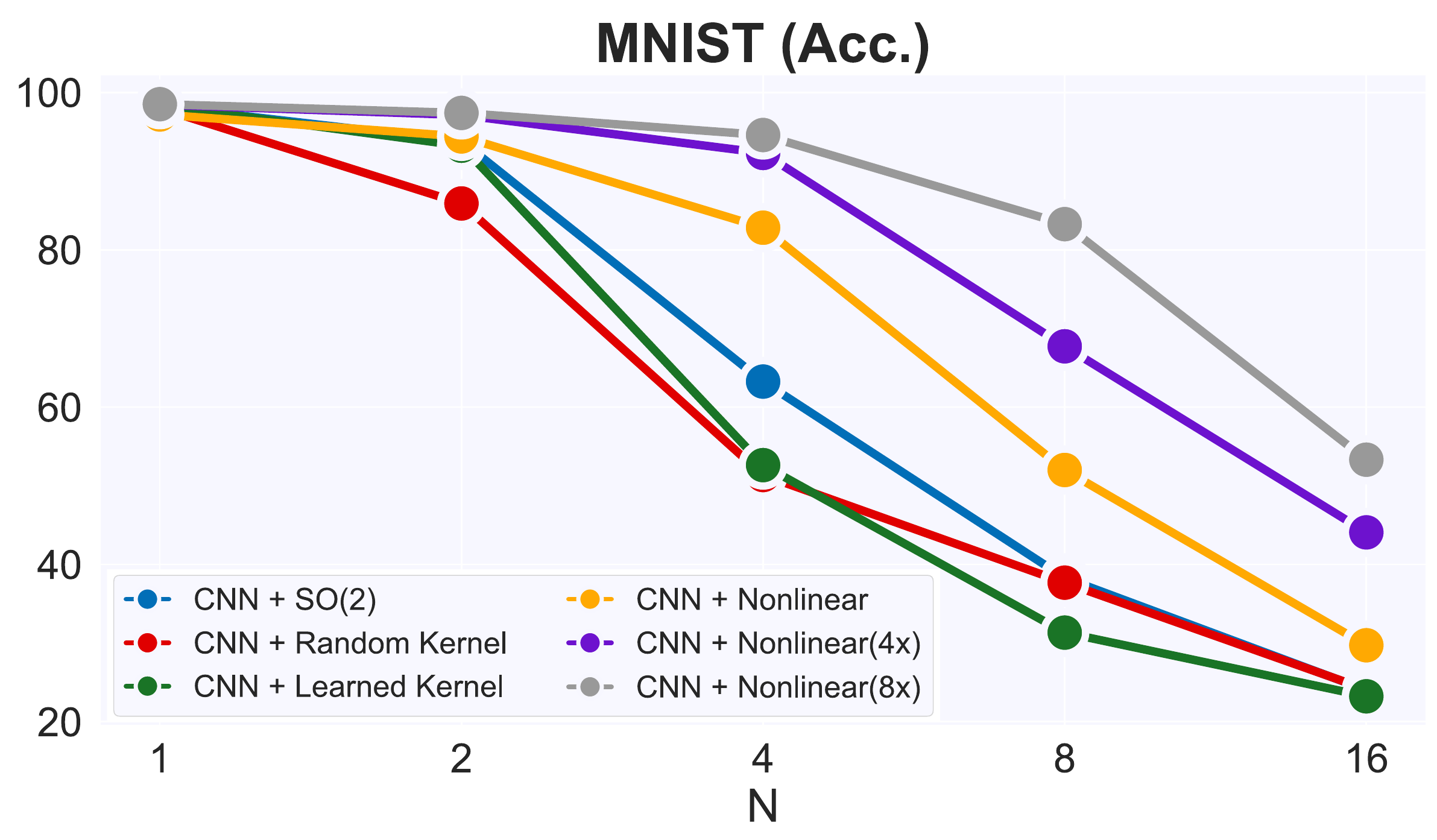}
    \caption{Averaged test accuracy for multiplexing CNN for $N=\{1, 2, 4, 8, 16\}$ inputs on the MNIST classification task, with different multiplexing strategies. Results are stable across multiple runs.}
    \label{fig:cnn-kers}
\end{figure}

Figure \ref{fig:cnn-kers} summarizes the performance of other separation functions we used for multiplexing CNN that can preserve spacial locality of images. During multiplexing, we can slide a 3x3 kernel over each input images before summing them up. This make CNN able to distinguish different inputs even with random initialized weights from $\mathcal N(0,1)$ (CNN+Random Kernel). We also make the weights of all $N$ multiplexing kernels learned but the difference is subtle (CNN+Learned Kernel), and the its performance is worse than a multiplexing CNN with 2D rotations. Also, we find that these multiplexing CNN can always answer at most two inputs correctly each time. This is because sliding a small kernel over image is a very constrained linear transformation that cannot do much to separate images, and the permutation symmetry has to be broken during the training dynamics to improve accuracy.

To increase the expressibility of separation functions while keep it compatible with CNN, we apply $N$ small convolutional nets with two layers of 16 3x3 kernels and tanh activation to input images, and sum up their activation maps (CNN+Nonlinear). A multiplexing CNN with this nonlinear separation function is similar to the MIMO approach in the previous study \cite{Rame21, Havasi21}, and we observe the performance changes consistent with the literature. When $N\leq 4$, its test average test accuracy is above 80\%, which is also better than multiplexing CNN with $\text{SO}(d)$. However, when $N > 4$, the performance drops rapidly. The accuracy across inputs becomes stabler with std. at about 7\% when $N\geq 8$. If we allow higher dimensionality of the multiplexed input over a single input, that is use $4$ activation maps (CNN+Nonlinear(4x)) or 8 activation maps (CNN+Nonlinear(8x)) instead of using a single activation map, we can keep improving the accuracy for larger $N$ while still providing improved throughput.

% Describe the setup mathematically
% Takeaways, more orthonormal with increasing N, model is forced to 
% 
\subsection{Memory overhead for multiplexed models}
\begin{figure}[ht]
    \centering
    \includegraphics[width=0.5\linewidth]{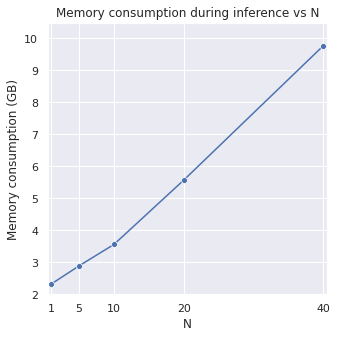}
    \caption{Memory overhead of multiplexed models during inference increases linearly with increasing N, with a very gentle slope (Memory overhead for N = 40 is $\sim4$x than N = 1).}
    \label{fig:memory_overhead}
\end{figure}
\vmn{We measure the memory overhead of various multiplexed models in Figure~\ref{fig:memory_overhead}. We use a fixed minibatch size of $60$ for all N and measure GPU memory during inference. We use the index demultiplexing strategy along with the Hadamard multiplexing strategy. We note that the memory increases linearly with increasing N as the number of inputs to the demultiplexing layers increases linearly with increasing N. However, the rate of growth is very gentle and the memory for N = 40 grows only by a mere $\sim4$x compared to the baseline model.}

\end{document}